%% file: main.tex
\documentclass[sigplan,nonacm, natbib=false]{acmart}
\renewcommand\footnotetextcopyrightpermission[1]{}
\renewcommand\authornotemark[1]{}

\usepackage{tikz}
\usepackage{amsmath}
\usepackage{colortbl}
\usepackage{graphicx}
\usepackage{epsfig,url,xcolor,xspace,balance}
\usepackage{enumitem}
\usepackage{tabularx}
\usepackage{multirow}
\usepackage{listings}
\usepackage{anyfontsize}
\usepackage{hyperref}
\usepackage{nccmath}
\hypersetup{
    colorlinks=true,
    citecolor=black,
    urlcolor=black, %
    linkcolor=black, %
}
\usepackage{algorithm}
\usepackage[noend]{algpseudocode}

\usepackage[style=numeric, maxbibnames=6]{biblatex}
\lstdefinestyle{codestyle}{
    commentstyle=\color{green},
    keywordstyle=\color{blue},
    numberstyle=\tiny\color{gray},
    stringstyle=\color{purple},
    basicstyle=\ttfamily\small,
    breakatwhitespace=false,         
    breaklines=true,                 
    captionpos=b,                    
    keepspaces=true,                 
    numbers=none,                    
    numbersep=5pt,                  
    showspaces=false,                
    showstringspaces=false,
    showtabs=false,                  
    tabsize=2,
    frame=single,
    escapechar={|}
}

\newcommand{\ignore}[1]{}

\newcommand{\derived}{transformed}

\newcommand{\tfdata}{\texttt{tf.data}\xspace}

\newcommand{\dataset}{\textsc{Dataset}\xspace}
\newcommand{\makeiterator}{\texttt{make\_iterator}\xspace}
\newcommand{\asgraphdef}{\texttt{serialize}\xspace}
\newcommand{\elementspec}{\texttt{element\_spec}\xspace}

\newcommand{\iterator}{\textsc{Iterator}\xspace}
\newcommand{\getnext}{\texttt{get\_next}\xspace}
\newcommand{\saveiterator}{\texttt{save}\xspace}
\newcommand{\restoreiterator}{\texttt{restore}\xspace}

\newcommand{\batchdataset}{\texttt{batch}\xspace}
\newcommand{\groupdataset}{\texttt{groupby}\xspace}
\newcommand{\cachedataset}{\texttt{cache}\xspace}
\newcommand{\concatenatedataset}{\texttt{concatenate}\xspace}
\newcommand{\fromfiledataset}{\texttt{from\_file}\xspace}
\newcommand{\frommemorydataset}{\texttt{from\_memory}\xspace}
\newcommand{\interleavedataset}{\texttt{interleave}\xspace}
\newcommand{\mapdataset}{\texttt{map}\xspace}
\newcommand{\prefetchdataset}{\texttt{prefetch}\xspace}
\newcommand{\repeatdataset}{\texttt{repeat}\xspace}
\newcommand{\shuffledataset}{\texttt{shuffle}\xspace}
\newcommand{\zipdataset}{\texttt{zip}\xspace}
\newcommand{\filterdataset}{\texttt{filter}\xspace}
\newcommand{\flatmapdataset}{\texttt{flat\_map}\xspace}
\newcommand{\unbatchdataset}{\texttt{unbatch}\xspace}

\newcommand{\reducedataset}{\texttt{reduce}\xspace}
\newcommand{\sharddataset}{\texttt{shard}\xspace}

\newcommand{\resnet}{\textsc{Resnet50}\xspace}
\newcommand{\maskrcnn}{\textsc{Mask-RCNN}\xspace}
\newcommand{\transformer}{\textsc{Transformer}\xspace}
\newcommand{\bert}{\textsc{BERT}\xspace}
\newcommand{\ssd}{\textsc{SSD}\xspace}
\newcommand{\gnmt}{\textsc{GNMT}\xspace}
\newcommand{\imagenet}{\textsc{ImageNet}\xspace}
\newcommand{\coco}{\textsc{COCO}\xspace}

\newcommand{\wmtg}{\textsc{WMT16}\xspace}
\newcommand{\wmtt}{\textsc{WMT17}\xspace}

\begin{document}

\date{}

\title[]{tf.data: A Machine Learning Data Processing Framework}

\author{Derek G.~Murray*}
\affiliation{%
  \institution{Microsoft}}
\thanks{*Work done while at Google}

\author{Ji\v{r}\'{i} \v{S}im\v{s}a}
\affiliation{%
  \institution{Google}}
  
\author{Ana Klimovic*}
\affiliation{%
  \institution{ETH Zurich}}

\author{Ihor Indyk}
\affiliation{%
  \institution{Google}}

\input{abstract}

\maketitle

\input{intro}

\input{requirements}
\input{design}

\input{eval}

\input{experience}

\input{related}

\input{conclusion}
\input{acknowledgements}

\renewcommand{\UrlFont}{\small\tt}
\printbibliography

\end{document}

%% file: abstract.tex
\begin{abstract}

Training machine learning models requires feeding input data for models to ingest. Input pipelines for machine learning jobs are often challenging to implement efficiently as they require reading large volumes of data, applying complex transformations, and transferring data to hardware accelerators while overlapping computation and communication to achieve optimal performance. We present \tfdata, a framework for building and executing efficient input pipelines for machine learning jobs. The \tfdata API provides operators which can be parameterized with user-defined computation, composed, and reused across different machine learning domains. These abstractions allow users to focus on the application logic of data processing, while \tfdata's runtime ensures that pipelines run efficiently.

We demonstrate that input pipeline performance is critical to the end-to-end training time of state-of-the-art machine learning models. \tfdata delivers the high performance required, while avoiding the need for manual tuning of performance knobs. We show that \tfdata features, such as parallelism, caching, static optimizations, and non-deterministic execution are essential for high performance. Finally, we characterize machine learning input pipelines for millions of jobs that ran in Google's fleet, showing that input data processing is highly diverse and consumes a significant fraction of job resources. Our analysis motivates future research directions, such as sharing computation across jobs and pushing data projection to the storage layer.

\end{abstract}

%% file: intro.tex
\section{Introduction}
\label{sec:intro}

\begin{figure}
\includegraphics[trim = 0.5cm 0cm 0.8cm 0.6cm,clip=true, width=\linewidth]
{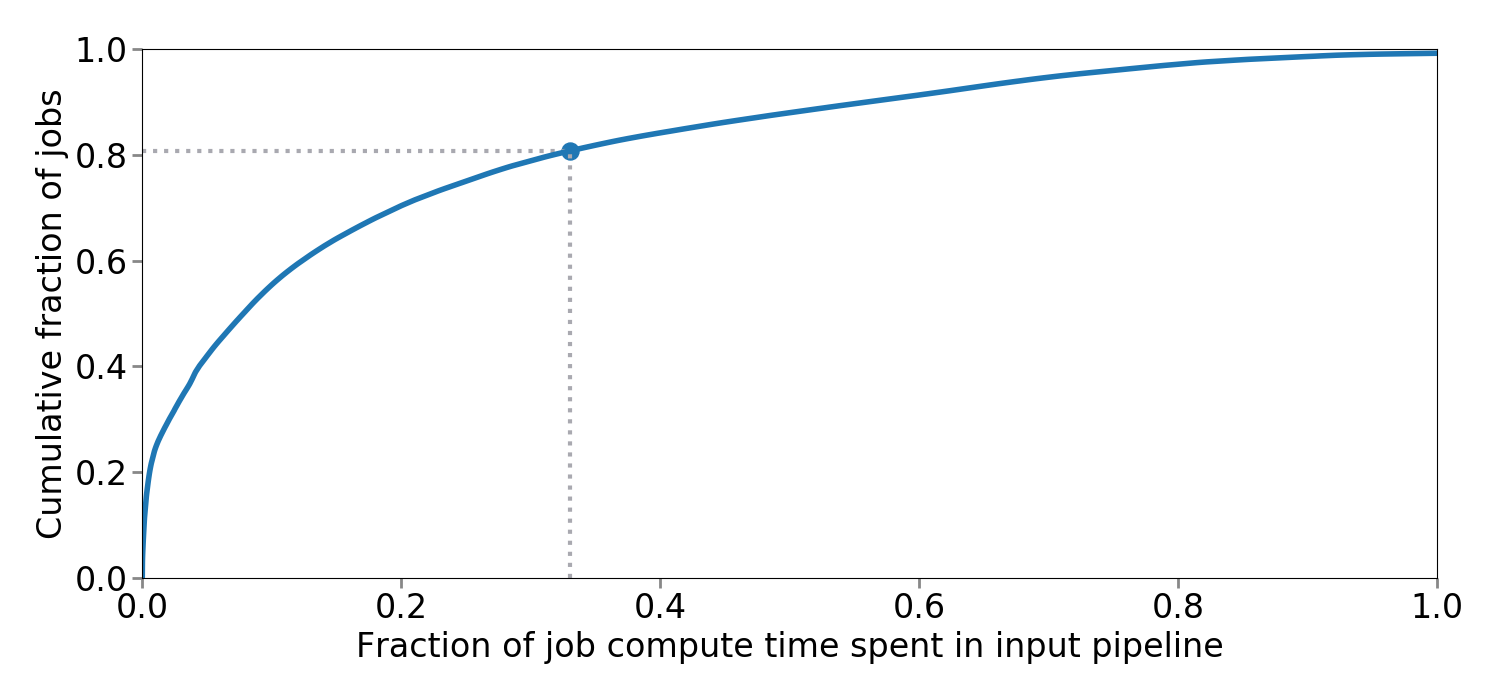}
\vspace{-0.7cm}
\caption{CDF showing the fraction of compute time that millions of ML training jobs executed in our fleet over one month spend in the input pipeline. 20\% of jobs spend more than a third of their compute time ingesting data.}
\label{fig:tfdata-gcu-fraction}
\vspace{-0.2cm}
\end{figure}

Data is the lifeblood of machine learning (ML). Training ML models requires steadily pumping examples for models to ingest and learn from. While  prior work has focused on optimizing the accuracy and speed of model training and serving, how we store and preprocess data for machine learning jobs has received significantly less attention. Across the millions of ML jobs we run in Google's datacenters every month, we observe that the input data pipeline accounts for significant resource usage and can greatly impact end-to-end performance. %
Figure~\ref{fig:tfdata-gcu-fraction} shows how the fraction of compute time that jobs spend in the input pipeline varies, where we define \textit{compute time} as the time spent on a hardware resource -- such as a CPU or an accelerator core -- scaled by the compute capability of that resource.
The marked point shows that $20\%$ of jobs spend more than a third of their compute time in the input pipeline. When taking into account the total compute time from all jobs in our analysis (\S~\ref{sec:discussion}), we find that $30\%$ of the total compute time is spent ingesting data. A complementary study of ML model training with public datasets found that preprocessing data accounts for up to 65\% of epoch time~\cite{coordl}.
This shows that input data pipelines consume a significant fraction of ML job resources and are important to optimize.  %

Input pipelines of machine learning jobs are often challenging to implement efficiently as they typically need to ingest large volumes of data, apply complex transformations, overlap communication and computation, and shuffle and batch data with various data ordering guarantees. For example, some jobs require that each example is visited exactly once before any example is visited a second time during training. %
Moreover, to achieve good performance and avoid input pipeline stalls, the data preprocessing should leverage parallelism and pipelining to overlap preprocessing with model training computations. Determining the optimal degree of parallelism and amount of data to prefetch is often challenging as it depends on the nature of the workload and the hardware resources available. %

Hardware accelerators used for ML training further increase the need for efficient input pipelines.
Today's accelerators, such as GPUs and TPUs, are tailored towards executing the linear algebra operations that are common in ML computations, but have limited support for common data preprocessing operations. Hence, input data is commonly processed on the CPU and feeding an accelerator with data at a sufficient rate to saturate its compute capabilities is becoming increasingly challenging. The high cost of accelerators compared to their CPU hosts makes it particularly important to ensure that accelerators operate at high utilization~\cite{google-cloud-pricing, aws-pricing}.

We present \tfdata, an API and a runtime for building and executing efficient input data pipelines for machine learning jobs. The \tfdata API provides generic operators that can be parameterized by user-defined functions, composed, and reused across ML domains. Inspired by the programming models of relational databases~\cite{volcano,db2}, declarative collection libraries~\cite{linq,java-streams}, and data-parallel big-data systems~\cite{dryadlinq,spark}, the \tfdata API consists of stateless \textit{datasets}, which are an abstraction for users to define their input pipeline, and stateful \textit{iterators}, which produce a sequence of elements and maintain the current position within a dataset. These abstractions allow users to focus on the application logic of their input pipeline and leave the task of executing the pipeline efficiently to the \tfdata runtime. In particular, \tfdata internally represents an input pipeline dataset as a graph and applies static optimizations using graph rewrites. Furthermore, \tfdata can automatically tune parameters such as the degree of parallelism and data prefetch buffer sizes, which are critical for performance yet often challenging for an average ML user to tune by hand. %

Our evaluation demonstrates that 1) input pipeline performance is critical to end-to-end training time of state-of-the-art ML benchmarks, 2) \tfdata is capable of improving input pipeline latency through a combination of software pipelining, parallelization, and static optimizations, and 3) \tfdata dynamic optimizations avoid the need to manually tune performance knobs. For example, we show that introducing parallelism and software pipelining to the input pipeline of a \resnet model training on the \imagenet dataset results in a $10.4\times$ decrease in time to convergence. Applying further optimizations with \tfdata, such as caching and static optimizations, improves training time by an additional $2\times$. We also demonstrate that \tfdata's auto-tuning matches the performance of expert hand-tuned input pipelines. %

The \tfdata API and runtime is open source and integrated in TensorFlow~\cite{tensorflow}. We have been using \tfdata in production since 2017 for a variety of ML training jobs, such as supervised learning, federated learning, and reinforcement learning; with different data modalities, including text, image, and video data. The system is currently used daily by hundreds of thousands of ML jobs in our fleet. 

We conduct a fleet-wide analysis of \tfdata jobs to characterize the input pipelines of millions of real machine learning jobs and identify opportunities for future work in data preprocessing systems. %
We find that the set of transformations applied in input pipelines varies greatly across jobs. For 75\% of jobs, the materialized dataset is smaller in size compared to the raw input data read from storage, which implies that preprocessing commonly decreases the volume of data. %
Most notably, we observe that identical input pipelines are  re-executed within and across jobs, suggesting that caching materialized datasets is a promising future direction to explore to improve the performance and efficiency of input data processing for ML. We motivate several other directions for future research based on our findings, such as processing data closer to storage and disaggregating input data processing from model training to avoid host resource bottlenecks.

%% file: requirements.tex
\section{Input Pipeline Requirements}
\label{sec:requirements}

\begin{figure}
\includegraphics[trim = 0.5cm 0.5cm 0.8cm 0.5cm,clip=true, width=0.95\linewidth]
{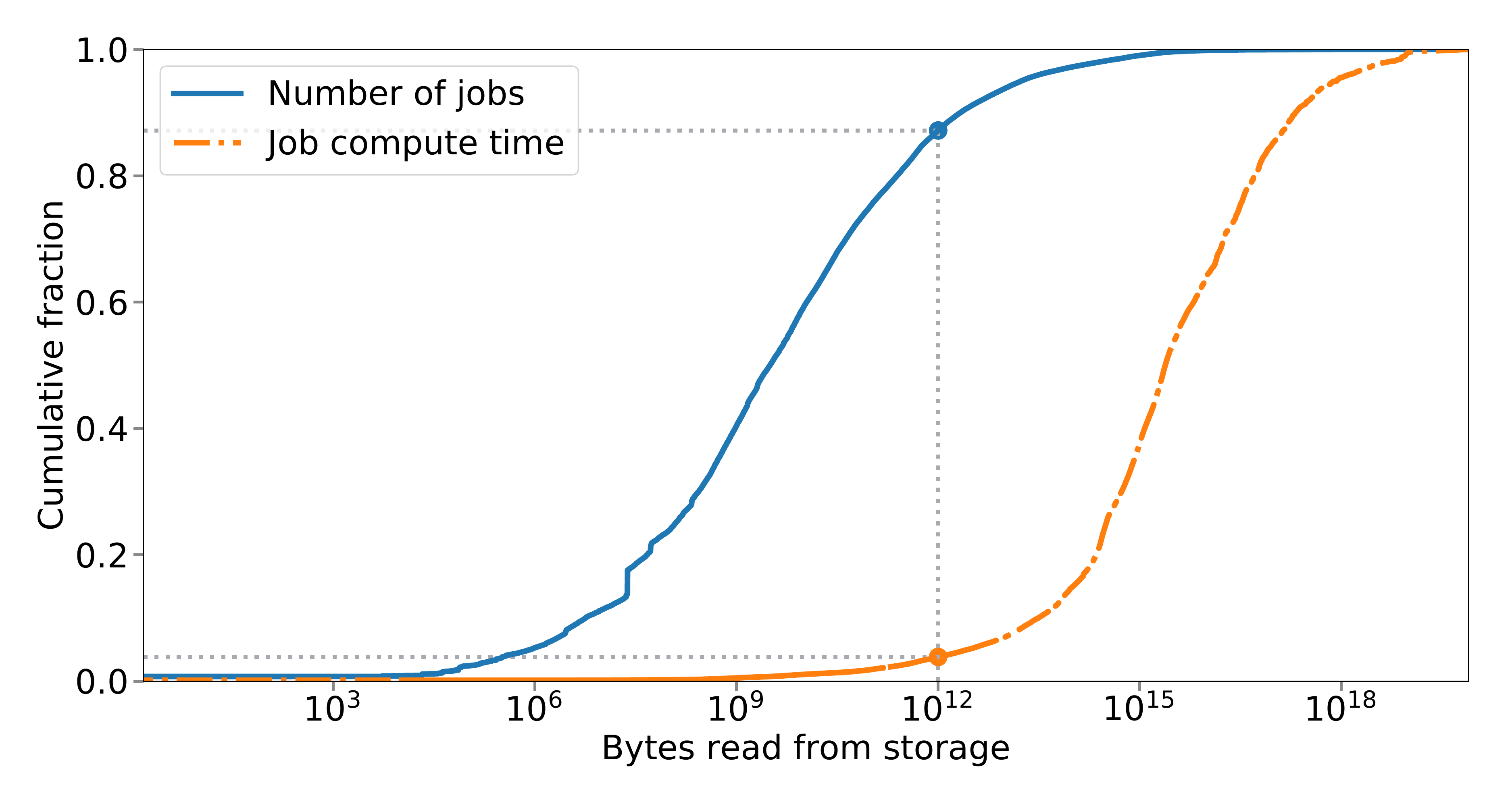}
\vspace{-0.4cm}
\caption{CDF of input data size across ML training jobs. 13\% of jobs read more than 1 TB of data. These jobs consume over 96\% of total compute resources.}
\vspace{-0.2cm}
\label{fig:dataset_size}
\end{figure}

Raw input data, such as images, audio, and text files, undergo both offline and online preprocessing before being ingested for model training.  Offline data preprocessing involves extracting features from raw data, validating data~\cite{data-validation-mlsys}, and converting data to binary formats, such as Avro~\cite{avro}, Parquet~\cite{parquet}, or TFRecord~\cite{tfrecord}, to enable higher throughput data ingestion. Batch computing frameworks such as Apache Spark~\cite{spark}, Beam~\cite{beam}, and Flume~\cite{flume} are commonly used for offline preprocessing. While some data transformations, such as normalization, are applied during offline preprocessing, ML training also requires applying transformations online as examples are fed to the model. For instance, image models commonly rely on data augmentation, e.g. randomly distorting images, to improve accuracy~\cite{imagenet, best-practices-cnns}. Such transformations multiply the size of the original dataset, making it prohibitive to store outputs in intermediate files. %
Our work focuses on online data preprocessing, which executes as part of the input pipeline of ML training jobs. 

The input pipeline of ML training can be characterized as a three-stage extract, transform, load (ETL) process. The first stage reads input data from a storage system. %
Machine learning jobs commonly train on large data sets.  Figure~\ref{fig:dataset_size} shows that $13$\% of jobs, out of the millions of jobs we analyzed, read at least $1$~TB of input data. This means that for a non-trivial fraction of training jobs, the input data cannot fit in memory. Furthermore, over $96$\% of total compute resources across jobs are spent in jobs that read over 1~TB of data.

The second stage transforms the data to a format amenable to ML training computation. It applies transformations to the input data, such as sampling, permuting, and filtering data to extract the subset of most relevant features. 
When training image models, it is common practice to apply data augmentation such as clipping, resizing, flipping, and blurring images. For text pipelines, training example commonly need to be grouped and batched based on sequence length.
Finally, the third stage loads the data onto the accelerator device that executes the training computation. 

ML training imposes unique requirements for input data pipelines. We describe these requirements below and summarize why they are not adequately addressed by other systems.

\paragraph{Data ordering} Unlike many data-parallel data processing platforms~\cite{mapreduce,dryadlinq,spark}, ML training is sensitive to the order in which records are delivered. The most common training algorithms are derived from \emph{stochastic} gradient descent~\cite{sgd}, which accesses the input examples pseudo-randomly. Empirically, convergence is more rapid when the algorithm makes multiple passes over input examples (called \emph{epochs}), and uses a different random permutation of the input data on each pass (or equivalently, samples examples without replacement within each epoch)~\cite{bottou-curiously-fast-convergence}. Furthermore, to improve system efficiency via vectorization and reduced communication, the input pipeline typically concatenates consecutive examples into a batch that is processed in a single training step.

The final parameters of a trained model can be sensitive to the exact order in which the input examples were consumed. To aid in debugging, especially when porting models between different hardware architectures, {\tfdata} must be able to produce random results in a deterministic order, according to a given seed. While such a feature is useful for debugging, it is in tension with high performance, since any variability in the element processing time could lead to head-of-line blocking. Therefore, while \tfdata defaults to deterministic execution, a user can disable it to mitigate the effect stragglers have on end-to-end performance.

Finally, both the end-to-end training computation and the individual epochs can take a long time to complete. To provide ordering guarantees in the presence of preemptions -- commonplace in our data centers -- the data processing computation for ML training jobs must be checkpointable.

\paragraph{Performance} A single training step consumes a batch of input elements and updates the current weights of the model. Often, the step computation runs on an accelerator device -- such as a GPU or TPU~\cite{tpuv2v3-cacm} -- that can compute vector floating point operations efficiently, although the computation may also run on a (multi-core) CPU. %
Ideally, the data processing computation is pipelined with the training computation, minimizing the likelihood that the training computation is blocked waiting for the next batch of elements and hence maximizing the utilization of valuable accelerator resources.

The input pipeline is responsible for fetching the raw input data from storage and transforming it into input features for the model. For example, the raw input for an image classification model might be a protocol buffer~\cite{protobuf} containing a JPEG-encoded image, and the input pipeline must convert the raw input into a dense three-dimensional array of floating point values corresponding to the RGB values of each pixel. Along the way, the input pipeline must extract and decode the JPEG and apply additional transformations such as affine transformations and colorspace changes to augment the training data~\cite{best-practices-cnns}. These activities are CPU-intensive, and must make efficient use of available CPU resources to maximize input pipeline throughput.

\paragraph{Ease of use} Machine learning workloads in a typical large organization span different domains, storage systems, data formats, and accelerator hardware. Therefore, it must be possible to combine pipeline stages in unanticipated ways, and extend the system with new data sources and transformations. To emphasize the importance of flexibility, in our fleet-wide analysis of ML jobs, we classified transformations into categories -- such as reading input data from storage, caching, batching, or shuffling -- and recorded the combination of transformation categories used by each job. While the $10$ most common combinations of transformations account for over $75$\% of jobs, there is a heavy tail with over $1000$ combinations of transformations in total.  %
In addition to supporting diverse input pipelines, we also require the input pipeline framework to address the tension between performance and ease-of-use. %
Optimizing an input pipeline can require expertise in how to structure operations and tune performance-related parameters, such as degrees of parallelism and pipeline buffer sizes. Hence, we require that {\tfdata} can optimize an input pipeline automatically. 

\smallskip
 Before designing {\tfdata}, we evaluated several existing input pipeline implementations, and found that they did not meet our requirements in one or more of the above areas: %
1) PyTorch's \texttt{DataLoader} API~\cite{pytorch-dataloader} is easy to use (it provides a simple Python interface), but its reliance on Python on the critical path -- despite the use of multiprocessing to work around the interpreter lock bottleneck -- and assumption of uniform random access to all input data, do not satisfy our performance requirement, especially for multi-terabyte datasets.
2) MXNet's \texttt{DataIter} API~\cite{mxnet_dataio} uses a native C++ implementation for greater performance than PyTorch, but it requires users to add native extensions in order to handle new preprocessing schemes. Therefore it does not help our users with diverse data processing needs, who tend to prefer programming in Python, and who are often restricted to memory-safe programming languages for security reasons.
3) NVIDIA's Data Loading Library (DALI) API~\cite{nvidia-dali} enables some preprocessing operations, such as image decoding, to be offloaded to a GPU. This offloading partially fulfils our performance requirement, but it lacks the flexibility to support heterogeneous preprocessing workloads and different types of accelerators.

In the next section, we present the \tfdata programming model, which is based on chaining higher-order functional transformations, and inspired by LINQ~\cite{linq}. Several data processing systems offer a similar programming model, including DryadLINQ~\cite{dryadlinq}, Spark~\cite{spark}, and Naiad~\cite{murray2013naiad}. We discuss them in more detail in \S~\ref{sec:relatedwork}. For pragmatic reasons, we did not consider using any of these systems, because the impedance mismatch with TensorFlow's C++ codebase would severely limit performance. Furthermore, these systems are designed to optimize data parallel computations, with a large number of independent values in each batch. This makes it difficult or inefficient for them to produce values sequentially, to fulfill the sequential ordering requirement. While one could use a system like Spark Streaming~\cite{spark-streaming} for online preprocessing and pass data to the ML framework through an in-memory buffer, the additional copies would have significant overhead due to the short step times in ML training workloads. In the training workloads we have analyzed, step times less than 1 ms are not uncommon and most workloads have step times less than 10ms. The extra copy overhead would be especially significant in the common case where memory bandwidth is the bottleneck. %

%% file: design.tex
\section{Design and Implementation}
\label{sec:design}

In \S~\ref{ss:design:datasets_and_iterators}, we present \tfdata's  API which enables users to compose and parameterize operators. In \S~\ref{ss:design:parallel} and \S~\ref{ss:design:auto-opt} we discuss key aspects of \tfdata's runtime.

\begin{table}[b]
    \centering
    \footnotesize
    \begin{tabular}{c|c}
        Method & Description \\
        \hline
        \makeiterator & Creates a new iterator over the dataset. \\
        \asgraphdef & Converts the dataset to a serialized expression. \\
        \elementspec & Returns the type signature of dataset elements. \\
    \end{tabular}
    \caption{\dataset interface}
    \label{tab:design:dataset_interface}
\end{table}

\subsection{Datasets and Iterators}\label{ss:design:datasets_and_iterators}

The {\tfdata} {\dataset} represents the stateless definition of an input pipeline as a (potentially infinite) sequence of elements. A dataset can either be a \textit{source dataset} that is created from primitive values (e.g.\ a matrix of floating-point numbers representing input examples, or a vector of strings representing filenames), or a \textit{{\derived} dataset} that transforms one or more input datasets into a new sequence of elements. The elements of a dataset are statically typed, and valid element types include tensors (with a specific element type and optional shape) and composite types (such as tuples, optionals, and nested datasets). Together, source and {\derived} datasets form an expression tree that represents the entire input pipeline. Table~\ref{tab:design:dataset_interface} shows the {\dataset} interface.

{\tfdata} includes source datasets that support common file formats and various {\derived} datasets which implement functional transformations and may be parameterized by user-defined functions (UDFs). The UDFs can be written in Python, and {\tfdata} uses TensorFlow's Autograph library to convert them into dataflow graphs~\cite{autograph}. Table~\ref{tab:design:datasets} summarizes the most common {\tfdata} transformations.

The {\tfdata} {\iterator} represents the current state of traversing a {\dataset}. An iterator provides sequential access to the elements of a dataset via the {\getnext} operation that either returns a typed element, or an error status such as "out-of-range" (EOF). In {\tfdata}, implementations of the {\iterator} interface are thread-safe, so multiple threads can call {\getnext} concurrently to improve throughput, at the expense of determinism. The interface also includes {\saveiterator} and {\restoreiterator} methods to support checkpointing.

The iterator interface (Table~\ref{tab:design:iterator_interface}) abstracts all details of how the elements are produced, including internal buffering and parallelism. Before applying optimizations, there is a one-to-one correspondence between dataset and iterator objects, but the optimizations in \S~\ref{ss:design:auto-opt} exploit the iterator abstraction to change the underlying dataset graph, and optimize how elements are produced, while presenting the same interface.

The example in Figure~\ref{fig:sequential-pipeline} illustrates a training loop that uses a \tfdata input pipeline to read elements from files, apply user-defined processing logic on each element and combine the processed elements into a mini-batch.

\begin{table}[]
    \centering
    \footnotesize
    \begin{tabular}{c|c}
        Dataset & Description \\
        \hline
        \batchdataset & Concatenates multiple elements into a single element. \\
        \cachedataset & Stores the input data in memory. \\ 
        \concatenatedataset & Concatenates two datasets. \\ 
        \fromfiledataset & Reads elements from a file. \\
        \frommemorydataset & Creates a singleton dataset from data in memory. \\
        \filterdataset & Returns elements matching a predicate. \\
        \flatmapdataset & Maps elements to datasets and flattens the result. \\
        \interleavedataset & Like \flatmapdataset, but mixes outputs from input elements. \\
        \mapdataset & Transforms individual elements. \\
        \prefetchdataset & Adds a buffer to pipeline input production. \\
        \reducedataset & Reduces a dataset to a single element. \\
        \repeatdataset & Produces the input dataset multiple times. \\
        \sharddataset & Selects a subset of elements from the dataset. \\
        \shuffledataset & Randomizes the order of elements. \\
        \unbatchdataset & Splits input elements on the 0th dimension. \\
        \zipdataset & Combines elements of multiple datasets into tuples. \\
    \end{tabular}
    \caption{Common {\tfdata} source and {\derived} datasets.}
    \vspace{-10pt}
    \label{tab:design:datasets}
\end{table}

\begin{table}[b]
    \centering  
    \footnotesize
    \begin{tabular}{c|c}
        Method & Description \\
        \hline
        \getnext & Returns the next element, or raises EOF. \\
        \saveiterator & Writes the iterator state to a file. \\
        \restoreiterator & Reads the iterator state from a file. \\
    \end{tabular}
    \caption{\iterator interface}
    \label{tab:design:iterator_interface}
\end{table}

\subsection{Parallel and Distributed Execution}\label{ss:design:parallel}

To efficiently utilize available host resources, \tfdata provides transformations that enable software pipelining, and parallel execution of computation and I/O. The \prefetchdataset transformation decouples the producer and consumer of data using an internal buffer, making it possible to overlap their computation. Input pipelines can use this transformation to overlap host computation, host-to-device transfer, and device computation.
The \mapdataset transformation takes an optional argument that specifies the degree of parallelism to use for applying the user-defined computation to input elements concurrently. The \interleavedataset transformation provides a similar optional argument that specifies the degree of parallelism to use for fetching data from input elements concurrently. In particular, the \interleavedataset transformation can parallelize I/O by interleaving data read from multiple files. By default, \tfdata transformations produce elements in a deterministic order. However, as deterministic ordering can lead to head-of-line blocking, the parallel \mapdataset and \interleavedataset transformations provide a mechanism for enabling non-deterministic ordering, which can result in better performance at the expense of reproducibility.

To illustrate the benefits of the above transformations, we revisit the example presented in Figure~\ref{fig:sequential-pipeline}. Let us assume that it takes $5$ms to read an element from the file, $2$ms to apply the user-defined logic to an element, and $1$ms to batch 10 elements. The accelerator would be idle for $(5 + 2) * 10 + 1 = 71$ms at the start of each iteration before data for the training computation becomes available.

\begin{figure}
\begin{lstlisting}[language=Python]
  ds = tf.data.|\color{green!40!black}from\_file|(["foo", ...])
  ds = ds.|\color{green!40!black}map|(parse).batch(batch_size=10)
  for elem in ds:
    train_step(elem)
\end{lstlisting}
\caption{Example of a training loop using \tfdata input pipeline. \texttt{parse} is a user-defined function for data processing.}
\label{fig:sequential-pipeline}
\end{figure}

\begin{figure}
\begin{lstlisting}[language=Python]
  ds = tf.data.|\color{green!40!black}from\_memory|(["foo", ...])
  ds = ds.|\color{green!40!black}interleave|(
         tf.data.|\color{green!40!black}from\_file|, cycle_length=2,
         num_parallel_calls=2)
  ds = ds.|\color{green!40!black}map|(parse, num_parallel_calls=10)
  ds = ds.|\color{green!40!black}batch|(batch_size=10)
  ds = ds.|\color{green!40!black}prefetch|(buffer_size=1)
  for elem in ds:
    train_step(elem)
\end{lstlisting}
\caption{Example of a training loop with \tfdata input pipeline that employs parallelism and software pipelining.}
\label{fig:parallel-pipeline}
\end{figure}

The \tfdata input pipeline in Figure~\ref{fig:parallel-pipeline} is semantically equivalent to that of Figure~\ref{fig:sequential-pipeline}. However, it uses 1) the optional \texttt{num\_parallel\_calls} argument of \interleavedataset and \mapdataset to parallelize I/O and computation respectively, and 2) \prefetchdataset to overlap the input pipeline computation with the training computation. As a result, the input pipeline in Figure~\ref{fig:parallel-pipeline} will take $max(10 * 5 / 2, 10 * 2 / 10, 1) = 25$ms to produce a batch (assuming a sufficiently slow consumer) and the input pipeline computation (of the next batch) will be overlapped with the training computation on the accelerator (for the current batch). If the training computation takes more than $25$ms, the data for each iteration of the training loop will be ready by the time the iteration starts. In \S~\ref{ss:design:auto-opt:dynamic} we describe a mechanism for auto-tuning parallelism and buffer sizes so that users do not have to tune them manually.

While \interleavedataset is typically used to parallelize I/O, it can also be used for parallel execution of multiple copies of an arbitrary input pipeline (operating over different shards of the input data). We have found this mechanism useful to speed up input pipelines bottlenecked by inherently sequential transformations, such as \filterdataset or \unbatchdataset.

In addition to supporting efficient single-host execution, we also designed \tfdata for distributed ML training computation use-cases, such as data parallel synchronous training, across multiple hosts (and accelerators per host). In this setup, each host has a \tfdata input pipeline providing data for the accelerators attached to the host. To provide for clean separation of epochs, the input data can be sharded across multiple files and the \sharddataset transformation ensures that different hosts operate over different shards of the data. The sharded input pipelines do not communicate with each other.

\subsection{Automatic Optimization}\label{ss:design:auto-opt}

{\tfdata}'s functional programming model enables it to provide multiple different implementations for a single input pipeline.  Automatic \emph{static}~(\S\ref{ss:design:auto-opt:static}) and \emph{dynamic}~(\S\ref{ss:design:auto-opt:dynamic}) optimizations improve {\tfdata}'s  performance and usability.

\subsubsection{Static Optimizations}\label{ss:design:auto-opt:static}

At run-time, {\tfdata} can reflect on the expression tree of any dataset and replace it with a more efficient version. We implemented static optimizations as a virtual dataset transformation that converts the input dataset to an expression tree, applies a suite of rewriting rules, and then evaluates the rewritten expression tree to produce an output dataset. The current implementation uses TensorFlow's \textsc{GraphDef} protocol buffer as the representation and the Grappler optimization framework~\cite{grappler} to manipulate these expression trees. We are investigating the use of MLIR~\cite{mlir} as a richer representation that will enable us to reuse optimizations from other domains.

As we gained experience with {\tfdata}, we created several custom transformation that fuse commonly adjacent transformations for performance reasons: \mapdataset + \batchdataset fusion, \shuffledataset + \repeatdataset fusion, \mapdataset + \mapdataset fusion, \mapdataset + \filterdataset fusion, and \filterdataset + \filterdataset fusion.  For example, the \mapdataset + \batchdataset fusion transforms $\mathrm{d}.\mathrm{map}(f).\mathrm{batch}(b)$ into $\mathrm{map\_and\_batch}(f, b)$, which is functionally equivalent but the implementation of the fused operator parallelizes and pipelines the copies of each element into the output batch with the processing of other batch elements. Many of the fusion optimizations in {\tfdata} are inspired by deforestation in functional languages~\cite{deforestation}. As the simplest example, the \mapdataset + \mapdataset fusion transforms $\mathrm{d}.\mathrm{map}(f).\mathrm{map}(g)$ expression into $\mathrm{d}.\mathrm{map}(g \circ f)$. This eliminates the per-element overhead of an iterator---a virtual call to {\getnext} and one of two function dispatches---and the composition $g \circ f$ may be optimized further by Grappler's standard optimization passes, such as arithmetic optimization and dead code elimination.

\tfdata static optimizations are not limited to fusions. The \textit{map vectorization} is a more advanced optimization that transforms $\mathrm{d}.\mathrm{map}(f).\mathrm{batch}(b)$ into $\mathrm{d}.\mathrm{batch}(b).\mathrm{map}(\mathrm{pfor}(f))$. In the transformed expression, $\mathrm{pfor}(f)$ applies $f$ to every slice of the batch in parallel~\cite{pfor}. This increases the efficiency of the resulting code by converting multiple invocations of a per-element operation (e.g. \texttt{tf.matmul()} into a single invocation of a batched operation (e.g. \texttt{tf.batch\_matmul()}) that itself has an efficient vectorized implementation. It also reduces the framework-induced overhead by replacing $b$ function invocations with a single invocation.   

\subsubsection{Dynamic Optimizations}\label{ss:design:auto-opt:dynamic}

In many cases, the optimal configuration for a {\tfdata} pipeline depends on properties of the input data (e.g.\ raw image sizes) and the available resources (e.g.\ number of CPU cores, RAM, and network bandwidth). Hence, {\tfdata} provides configuration parameters such as the degree of parallelism for \mapdataset transformations and the size of the buffer for the \prefetchdataset transformation. %

To avoid the need for users to manually tune performance-related knobs, the {\tfdata} runtime contains an \textit{auto-tuning} mechanism that allocates CPU and RAM resources across various parts of the input pipeline in a way that minimizes the (expected) latency of the input pipeline producing an element. In the rest of this section, we refer to the time it takes for an iterator to produce an element as its \textit{output latency} and the output latency of an input pipeline is the output latency of the iterator for its final transformation. 

To perform auto-tuning, {\tfdata} executes the input pipeline in a light-weight harness, which maintains a tree representation of the iterators currently executing as part of the input pipeline and measures the processing time spent in each of the iterators. The root of the tree is the iterator producing data for training computation, the leaves of the tree correspond to source dataset iterators, and edges are implied by the input-output relationship between {\derived} datasets' iterators and their inputs. The tree structure can change over time as transformations such as \interleavedataset or \repeatdataset create multiple iterators during their lifetime.

The auto-tuning implementation uses the processing time and the input pipeline structure to build an analytical model that is used to estimate how input pipeline parameters affect end-to-end latency. The estimating function is a composition of the output latencies of individual iterators as functions of tunable parameters, iterator's processing time and inputs' output latency. The outermost function of the composition is the one for the final iterator. For synchronous transformations (i.e. transformations that do not decouple producer and consumer), the output latency of an iterator is a linear function of the output latencies of its inputs and the processing time spent in the iterator. For asynchronous transformations, such as \prefetchdataset and the parallel \mapdataset and \interleavedataset, the output latency of an iterator is no longer linear and additionally depends on the parallelism, buffer size, and the rate of the consumer. In particular, the expected output latency of the iterator is computed as the output latency of its input(s) multiplied by the probability that the buffer is empty, which we model using an M/M/1/k queue~\cite{mm1k} and estimate as:
\useshortskip
\begin{equation}
\large
p_{empty} =
\begin{cases}
    \frac{1}{n + 1}                              & {\normalsize \text{if $x = y$}}\\ 
    \\
    \frac{1 - \frac{x}{y}}{1 - \left(\frac{x}{y}\right)^{n+1}} & \text{\normalsize otherwise}
\end{cases}
\label{eq:buffer_empty}
\end{equation}
\useshortskip
\noindent where $n$ is the buffer size, $x$ is the producer rate, computed from the output latency of the iterator input(s), and $y$ is the consumer rate, computed from the frequency of \getnext calls. Note that the producer rate, $x$, in general depends on upstream computation, while the consumer rate, $y$, in general depends on downstream computation. We traverse the iterator tree depth first to estimate both $x$ and $y$ in a single traversal.

\begin{figure}[htbp]
\hspace*{-1cm}
\centering
\scalebox{0.85}{%
\begin{tikzpicture}
  [
    auto,
    block/.style    = { rectangle, draw=blue, thick, 
                        fill=blue!20, text width=7em, text centered, rounded corners, minimum height=1.2em },
  ]
  \node [text width=4cm,text centered] (model) at (2.7, 10) {pipeline};
  \node (trainer) at (2.7, 9.3) {data consumer};
  \node [block] (prefetch) at (2.7, 8.3) {\prefetchdataset \\ buffer size $=2$};
  \node [block] (batch) at (2.7, 6.8) {\batchdataset \\ batch size $=10$};
  \node [block] (map) at (2.7, 4.9) {\mapdataset \\ parallelism $=5$ \\ buffer size $=5$};
  \node [block] (interleave) at (2.7, 2.5) {\interleavedataset \\ parallelism $=1$ \\ buffer size $=1$ \\ cycle length $=2$};
  \node [block] (fromfile) at (2.7, 1) {\fromfiledataset};
  \foreach \from/\to in {trainer/prefetch,prefetch/batch,batch/map,map/interleave,interleave/fromfile}
    \draw [-, line width=0.5mm] (\from) edge (\to);

  \node [text width=2.5cm,text centered] (consumer_rate) at (0, 10) {consumer rate, \getnext calls per second};
  \node (prefetch_cr) at (0, 8.3) {$\frac{1}{0.01}=100$};
  \node (batch_cr) at (0, 6.8) {$100$};
  \node [text width=4cm,text centered] (map_cr) at (0, 4.9) {$100*\text{batch size}$ \\ $=1000$};
  \node (interleave_cr) at (0, 2.5) {$1000$};
  \node (fromfile_cr) at (0, 1) {$\frac{1000}{\text{cycle length}}=500$};
  \foreach \from/\to in {prefetch_cr/batch_cr,batch_cr/map_cr,map_cr/interleave_cr,interleave_cr/fromfile_cr}
    \draw [->, line width=0.5mm] (\from) edge (\to);
  
  \node [text width=4cm,text centered] (output_latency) at (5.8, 10) {output \\ latency, ms};
  \node [text width=4cm,text centered] (prefetch_ol) at (5.8, 8.3) {$x=\frac{1000}{36.5}=27.4,$ \\ $y=100,\, n=2$ \\ $36.5* p_{empty}=27$};
  \node [text width=4cm,text centered] (batch_ol) at (5.8, 6.8) {$10 * 3.55 + 1=36.5$};
  \node [text width=4cm,text centered] (map_ol) at (5.8, 4.9) {producer time $=4.15+\frac25 = 4.55$ \\ $x=\frac{1000}{4.55}=220,$ \\ $y=1000, \,n=5,$ \\ $4.55* p_{empty}=3.55$};
  \node [text width=4cm,text centered] (interleave_ol) at (5.8, 2.5) {$x=\frac{1000}{5}=200,$ \\ $y=1000, \,n=1,$ $5* p_{empty} = 4.15$};
  \node (fromfile_ol) at (5.8, 1) {$5$};
  \foreach \from/\to in {prefetch_ol/batch_ol,batch_ol/map_ol,map_ol/interleave_ol,interleave_ol/fromfile_ol}
    \draw [<-, line width=0.5mm] (\from) edge (\to);
\end{tikzpicture}
}
\caption{Output latency estimation: the downward traversal computes the consumer rate starting with the root adjusting it by the number of concurrent \getnext calls from the consumer and the number of iterators. The upward traversal can compute the output latency of each iterator in the tree since by the time the traversal returns to an iterator, the output latency of its inputs is known. Asynchronous transformations \prefetchdataset, parallel \mapdataset and \interleavedataset use \eqref{eq:buffer_empty} to estimate the output latency, whereas a synchronous \batchdataset produces an estimate with a linear function of its own processing time and output latency of its input.}
\label{fig:output-latency}
\end{figure}
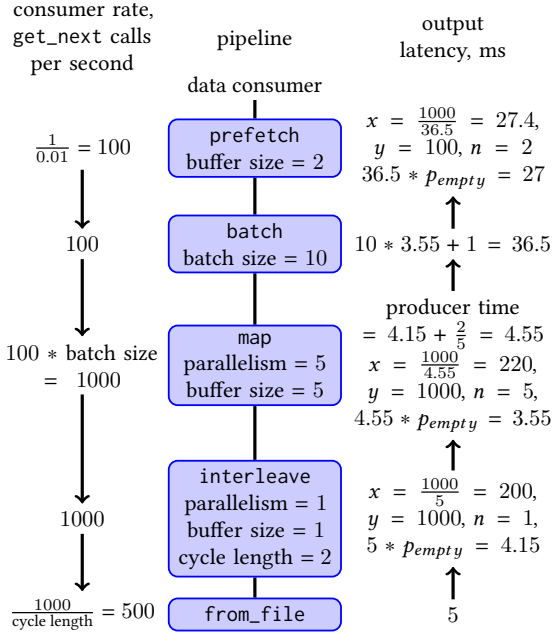

To illustrate how the estimation works, let's revisit the example from Figure~\ref{fig:parallel-pipeline}, additionally assuming that 1) the \texttt{num\_parallel\_calls} and \texttt{buffer\_size} arguments are set to the special \texttt{AUTOTUNE} value to enable auto-tuning, 2) the training computation requests data every $10$ms on average, and 3) the auto-tuning harness is estimating the following combination: \interleavedataset parallelism $1$ and buffer size $1$, \mapdataset parallelism $5$ and buffer size $5$, and \prefetchdataset buffer size $2$. Figure~\ref{fig:output-latency} gives an example of how \tfdata computes the output latency for such a pipeline.

\tfdata creates a background thread that periodically uses the estimation process above to evaluate different combinations of parallelism and buffer sizes for tunable transformations. Parameters are chosen to minimize the expected output latency of the input pipeline subject to CPU and RAM budget constraints. The optimization uses a gradient descent algorithm and is depicted in Figure~\ref{fig:autotuning-optimization}. The optimization period ranges from milliseconds to seconds and is determined automatically based on changes to the input pipeline structure and execution time.
 
\begin{figure} []
\begin{lstlisting}[language=Python]
while True:
  model = pipeline.get_analytical_model()
  params = model.get_tunable_parameters()
  best_latency = INFINITY
  latency = model.latency()
  while (best_latency - latency >= EPS and 
         model.resource_usage() <= BUDGET):
    best_latency = latency     
    params -= DELTA * model.latency_grad()
    latency = model.latency()
  pipeline.set_tunable_parameters(params)
  sleep(OPTIMIZATION_PERIOD)
\end{lstlisting}
\vspace{-10pt}
\caption{Periodic optimization of tunable parameters.}
\label{fig:autotuning-optimization}
\end{figure}

An important aspect of the optimization is its ability to minimize output latency of the end-to-end input pipeline as opposed to minimizing the output latency of individual transformations. As different transformations share the same CPU and RAM resources, locally optimal decisions may lead to excessive parallelism and buffering, which in turn lead to inefficient thread scheduling and poor cache locality, negatively affecting end-to-end performance.

 The ability to perform the optimization analytically is essential; it allows \tfdata to quickly find a good configuration without affecting the performance of the real input pipeline while evaluating sub-optimal configurations. Once the background thread identifies a configuration to use, it updates the parallelism and buffer sizes of the actual input pipeline accordingly. For most input pipelines the optimization takes microseconds to milliseconds to complete.

%% file: eval.tex
\section{Evaluation}
\label{sec:eval}

To evaluate \tfdata we seek to answer the following questions: 1) how do \tfdata's performance-related features affect input pipeline throughput, 2) how do input pipeline optimizations impact the end-to-end time to reach a target accuracy when training state-of-the-art ML models, and 3) how does \tfdata performance compare to other systems.

For our evaluation, we used the open-source MLPerf~\cite{mlperf} benchmark suite, which is the de facto standard for evaluating ML software and hardware systems by measuring how fast a system can train models to a target quality metric. We use \tfdata to express and execute input pipeline computation in MLPerf benchmarks. Our evaluation considers the following combinations of model architectures and input data: 1) \resnet~\cite{resnet} with \imagenet~\cite{imagenet}, 2) \ssd~\cite{ssd} with \coco~\cite{coco},  3) \maskrcnn~\cite{maskrcnn} with \coco~\cite{coco}, 4) \gnmt~\cite{gnmt} with \wmtg~\cite{wmt16}, and 5) \transformer~\cite{transformer} with \wmtt~\cite{wmt17}. 

Table~\ref{tab:input-data} summarizes the attributes of the MLPerf datasets, which range from 135~MB to 140~GB in size. Though these public datasets fit in memory before decompression and/or data augmentations, in  Section~\ref{sec:experience:usage} we discuss our experience with production workloads which commonly preprocess larger-than-memory datasets (Figure~\ref{fig:dataset_size}). When dealing with such datasets, \tfdata's prefetching and software pipelining optimizations become even more critical for end-to-end performance.

\begingroup
  \setlength{\tabcolsep}{1pt} %
 \renewcommand{\arraystretch}{1} %
 \begin{table}[]
 \centering
 \small
 \vspace{2pt}
 \begin{tabular}{ p{2cm} p{3cm} p{2cm} p{1cm}}
  {\bf Dataset} & {\bf Domain} & {\bf Artifacts} & {\bf Size} \\
 \hline
 \textsc{{ImageNet}} & {image classification} & 1.3M images & 140GB \\
 \textsc{{COCO}} & {object detection} & 330K images & 19GB \\
  \textsc{{WMT16}} & {translation} & 4M pairs & 1.3GB \\
 \textsc{{WMT17}} & {translation} & 4.5M pairs & 720MB \\
 \hline
 \end{tabular}
 \caption{\label{tab:input-data} MLPerf input data overview.} 
 \vspace{-.2cm}
 \end{table}
\endgroup

\begin{table*}[]
\small
\begin{tabularx}{\linewidth}{*1{>{\raggedright\arraybackslash}X}   *8{>{\centering\arraybackslash}X}}
  & Parallel computation &  Parallel I/O  & Software pipelining & Non-deterministic & Caching & Static Optimization & No intra-op parallelism \\
\resnet      & \checkmark & \checkmark & \checkmark & \checkmark & \checkmark & \checkmark & \checkmark \\
\ssd         & \checkmark & \checkmark & \checkmark & \checkmark & \checkmark & \checkmark & \checkmark \\
\footnotesize{\maskrcnn} & \checkmark & \checkmark & \checkmark & \checkmark & \checkmark & \checkmark & \checkmark \\
\gnmt        & \checkmark & \checkmark & \checkmark & \checkmark & \checkmark & \checkmark &  \\
\transformer & \checkmark & \checkmark & \checkmark & \checkmark &            &             &            \\
\end{tabularx}
\caption{\tfdata features used by different MLPerf benchmarks.}
\label{tab:mlperf_optimization}
\vspace{-0.48cm}
\end{table*}

Table~\ref{tab:mlperf_optimization} shows the various performance-related features of \tfdata used in the input pipeline portion of our MLPerf benchmark implementations. All input pipelines used the \mapdataset, \interleavedataset, and \prefetchdataset transformations for parallel computation, parallel I/O, and software pipelining, respectively. Non-deterministic ordering was also used by all pipelines to mitigate the effect of stragglers. With the exception of \transformer, the input pipelines used static \tfdata optimizations to benefit from transformation fusion and the \cachedataset transformation to materialize intermediate preprocessing artifacts in memory to avoid their recomputation across epochs. Note that intermediate artifacts cannot always be materialized as they may be a result of a randomized transformation which produces a different result each epoch. Finally, the image-based input pipelines (\resnet, \ssd, and \maskrcnn) also disabled intra-op parallelism for \tfdata computation. Intra-op parallelism makes it possible to parallelize execution of individual TensorFlow ops, such as \texttt{tf.matmul}, but this comes at the expense of increased CPU usage. For \tfdata input pipelines, intra-op parallelism generally provides little benefit (as there is plenty of inter-op parallelism) and can actually hurt performance of input pipelines that fully utilize CPU resources.

\subsection{Input Pipeline Experiments}\label{ss:eval:input_pipeline}

\textbf{Methodology: }
To evaluate the effect of \tfdata performance-related features on input pipeline throughput, we executed the input pipeline portion of our MLPerf benchmark implementations in a tight loop (with no model training computation) and measured the time it takes to process an epoch's worth of data. We used a single machine with 56 Intel Xeon 2.60~GHz CPU cores, 128~GB of RAM, and the input data stored on a 1~TB Samsung SM961 SSD. We limited the \resnet experiment to only use $60\%$ of the \imagenet data to make sure that an epoch's worth of data can be cached in memory. For each of the input pipelines we ran the following experiments: 1) a baseline which does not use any \tfdata performance features (i.e. sequential reading and processing), 2) a version that uses expert-tuned~\footnote{Expert-tuned parallelism sets \mapdataset parallelism to the number of CPU cores available on the machine, \interleavedataset parallelism to a constant between 10 and 64 tuned based on available I/O bandwidth, and the \prefetchdataset buffer size to an empirically tuned multiple of batch size.} parallelism for I/O and compute, 3) a version that uses all \tfdata performance features in Table~\ref{tab:mlperf_optimization} with expert-tuned parallelism, and 4) a version that uses all \tfdata performance features with auto-tuned parallelism. Note that even though the baseline does not use input pipeline parallelism, TensorFlow may still parallelize the user-defined computation in \mapdataset.

\begin{figure}
\includegraphics[trim = 0.5cm 0.3cm 0.8cm 0.5cm,clip=true, width=\linewidth]
{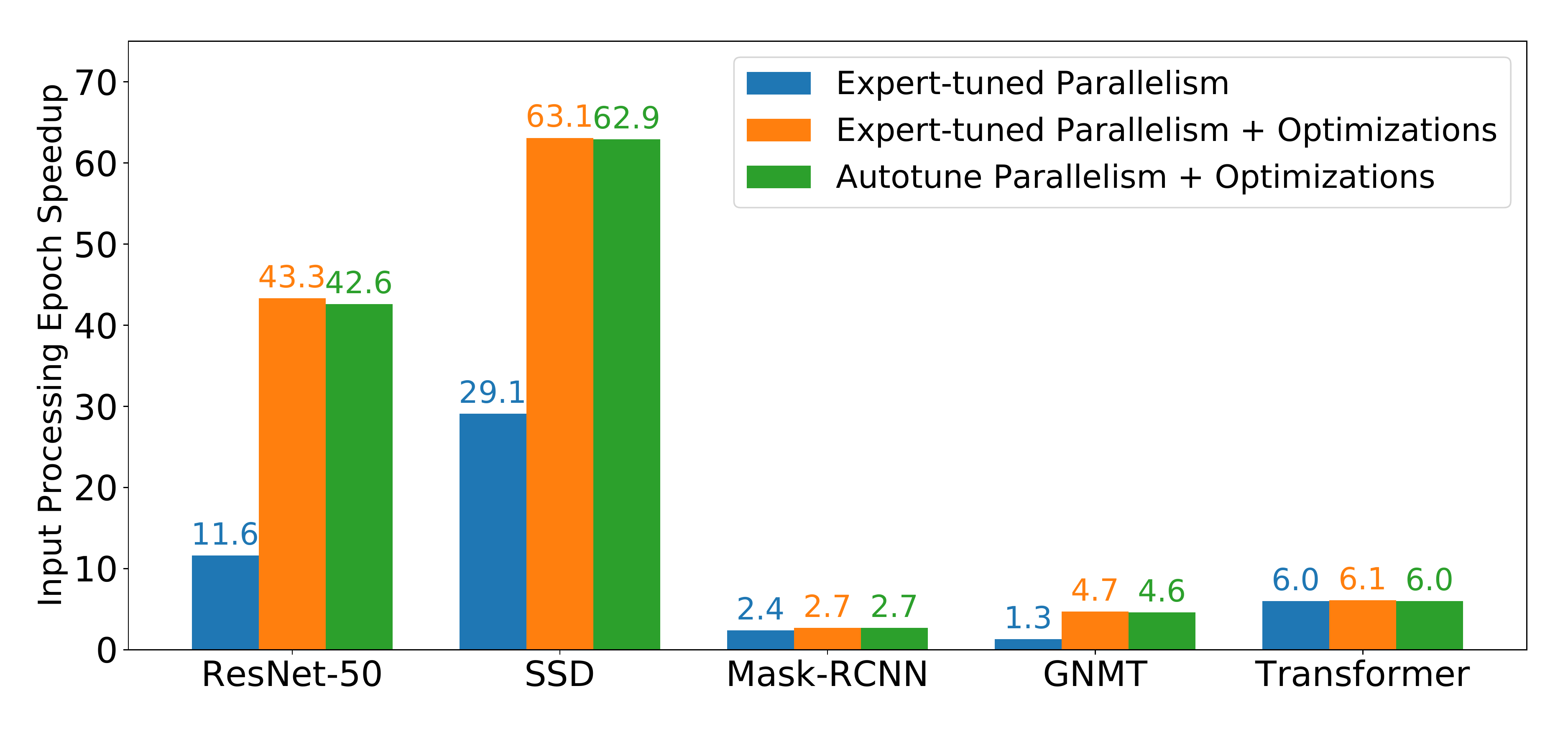}
\vspace{-0.6cm}
\caption{Speedup of input pipeline processing time with different configurations, relative to a sequential input pipeline.}
\label{fig:input_pipeline_evaluation}
\end{figure}

\textbf{Results:} Figure~\ref{fig:input_pipeline_evaluation} shows the mean duration of a single epoch, normalized to the epoch duration of the baseline, which does not use any \tfdata performance-related features. On the 56 core machine used for the experiment, the speedups ranged from $2.7 \times$ (\maskrcnn) to $63.1 \times$ (\ssd). Since we are parallelizing both compute and I/O it is possible to achieve speedup greater than $56 \times$.

The performance of \resnet and \ssd input pipelines benefits significantly from the application of \tfdata performance related features and the input pipeline can fully utilize the available CPU.  In particular, \mapdataset + \batchdataset fusion yields the most significant speedup among static optimizations for these two benchmarks, as it enables computing multiple batches in parallel. In contrast, the performance of \maskrcnn, \gnmt, and \transformer input pipelines benefits from the application of \tfdata performance-related features to a lesser extent. For \maskrcnn, the reason for the limited speedup is two-fold: 1) the baseline employs parallelism as the user-defined computation applied to each element can be parallelized by TensorFlow and 2) the input pipeline is bottlenecked by batching, which is performed sequentially because of an intermediate step between \mapdataset and \batchdataset in the pipeline that prevents  \mapdataset + \batchdataset fusion. %
Similarly, the text pipelines (\gnmt and \transformer) did not benefit from \mapdataset + \batchdataset fusion as elements need to be grouped based on size after the \mapdataset operation before they are batched, but the \tfdata runtime does not currently support \mapdataset+\groupdataset+\batchdataset fusion. Most benchmarks saw less than 4\% improvement in training time with non-deterministic vs. deterministic data ordering, however \resnet benefited more (approx. 40\% throughput improvement) as its dataset (\imagenet) has a wide distribution of image sizes, and non-deterministic ordering avoids head-of-line blocking.

For all of the input pipelines, using auto-tuned parallelism instead of expert hand-tuned parallelism results in comparable performance. This demonstrates that the algorithm described in \S~\ref{ss:design:auto-opt} is able to automatically configure performance knobs similar to a human expert.

\subsection{End-to-End Experiments}\label{sec:eval:large-scale}

\textbf{Methodology:} To evaluate how input pipeline performance optimizations with \tfdata translate to end-to-end performance benefits when training state-of-the-art ML models, we measured the time it takes to reach target accuracy with our \tfdata-based implementation of the MLPerf benchmarks. We executed each benchmark using 8 hosts with 112 CPU cores, 100~GB of RAM, and 8 TPUv3 accelerators each~\cite{tpuv2v3-cacm}. For the \maskrcnn benchmark, we used 400~GB RAM per host to ensure that intermediate artifacts can be fully cached in memory. We ran the following experiments for each benchmark: 1) a baseline that trains the MLPerf model with sequential reading and processing of input data, 2) a version that uses expert-tuned parallelism for I/O and compute in the input pipeline, 3) a version that uses all \tfdata performance features with expert-tuned parallelism, and 4) a version that uses all \tfdata performance features with auto-tuning.

\begin{figure}
\includegraphics[trim = 0.5cm 0.3cm 0.8cm 0.5cm,clip=true, width=\linewidth]
{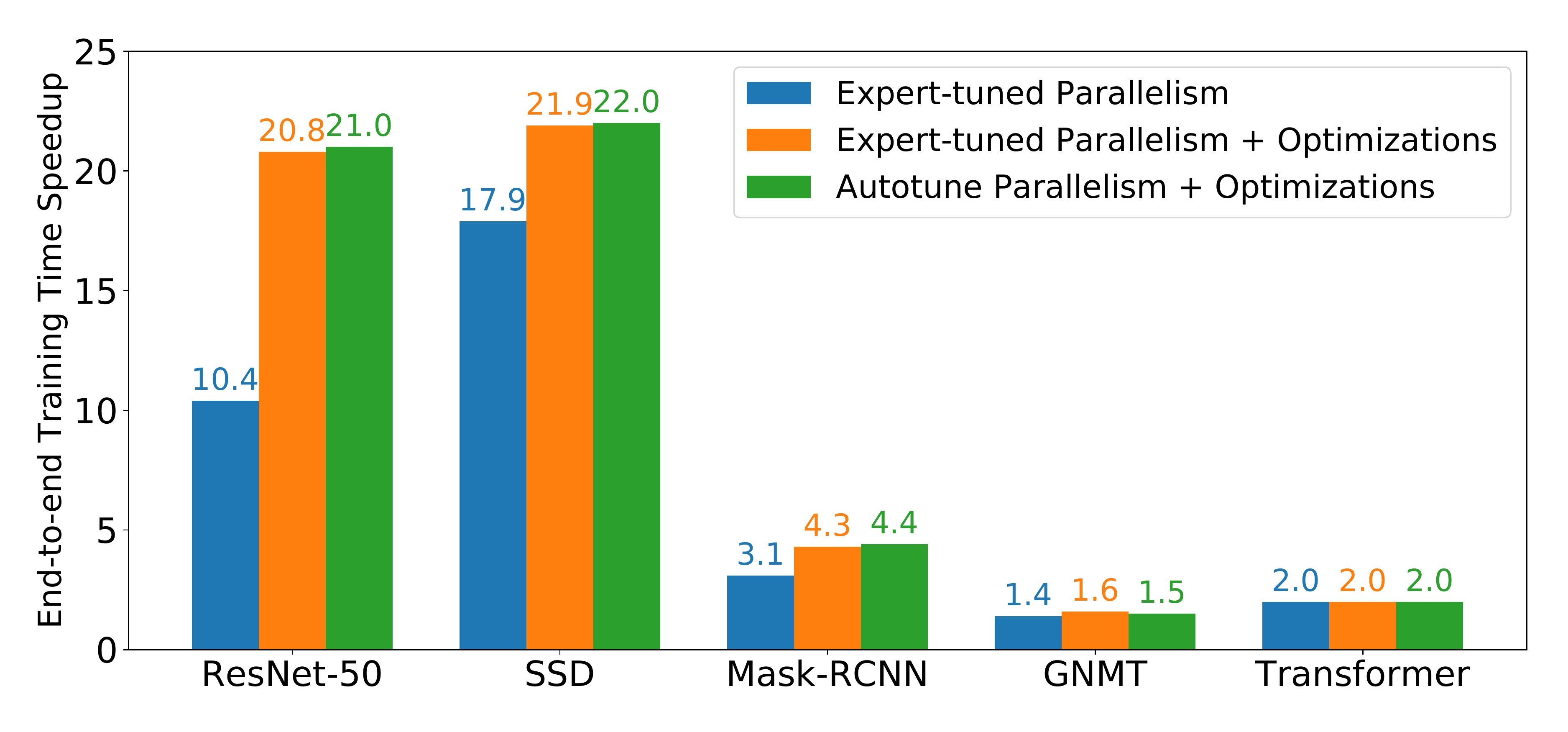}
\vspace{-0.8cm}
\caption{Speedup of the time to convergence for MLPerf workloads with \tfdata optimizations, relative to execution with a sequential input pipeline.} 
\label{fig:mlperf_speedup}
\vspace{-0.2cm}
\end{figure}

\textbf{Results:} Figure~\ref{fig:mlperf_speedup} shows the end-to-end training time speedup (relative to the model training time with a sequential input pipeline) for each MLPerf benchmark.  We draw several insights from these results. First and foremost, the performance of the input pipeline significantly affects the end-to-end training performance. Second, computation and I/O parallelism is necessary but not sufficient to match the rate at which accelerators perform training computation. Compared to using a sequential input pipeline as the baseline, adding software pipelining and parallelism in the input pipeline improved end-to-end training time by $7\times$ on average across the five MLPerf benchmarks.
For image-based input pipelines (\resnet, \ssd, and \maskrcnn), the end-to-end performance benefited further from the application of \tfdata performance-oriented features, providing an additional $2 \times$, $1.2 \times$, $1.4 \times$, speedup respectively. For text-based input pipelines (\gnmt and \transformer), parallelism and software pipelining alone were sufficient to match the rate at which data was consumed by the training computation.

Figure~\ref{fig:mlperf_speedup} also compares the training time with expert-tuned \tfdata configuration to training time with auto-tuned configuration. Similarly to the input pipeline experiments, we find that using \tfdata's dynamic optimizations to select parameters such as the degree of parallelism and prefetch buffer sizes leads to similar performance compared to the expert tuned pipelines. The end-to-end time to convergence with dynamic tuning is within $1\%$ of the time to convergence with expert tuned input pipelines for \resnet, \ssd, \maskrcnn, and \transformer and within $4\%$ for \gnmt. This demonstrates that \tfdata can effectively relieve users from the burden of hand-tuning input pipeline configurations.

Finally, we also verified that \tfdata optimizations enable input pipelines to match the rate at which accelerators perform training computations for state-of-the-art models. For each MLPerf benchmark, we measured the time it takes to ingest a batch of data and perform the model computation when using 1) an optimized \tfdata input pipeline versus 2) an artificial input pipeline that produces data as fast as possible (by caching a single batch and repeating it infinitely). The artificial pipeline does not perform any data processing and hence serves as an upper bound on input pipeline performance. Step times with optimized \tfdata pipelines match the upper-bound performance, hence the MLPerf benchmarks are no longer input bound after \tfdata optimizations.

\subsection{Comparison to Other Systems}

\begin{table}[t]
    \centering
    \small
    \begin{tabular}{c l c}
        {\bf Input data framework} & {\bf Hardware}  & {\bf Epoch duration (s)} \\
        \hline
        PyTorch DataLoader & CPU-only & 213 \\
        NVIDIA DALI & CPU-only & 777 \\
        NVIDIA DALI & CPU + 1 GPU  & 172 \\
        NVIDIA DALI & CPU + 2 GPUs & 107 \\
        \textbf{\tfdata} & CPU-only & \textbf{110} \\
    \end{tabular}
    \vspace{0.17cm}
    \caption{\imagenet-\resnet input data processing time with \tfdata vs. NVIDIA DALI and PyTorch DataLoader.} %
    \label{tab:eval:resnet}
    \vspace{-5pt}
\end{table}

To evaluate how \tfdata compares to other ML input data processing systems, we implement a standard \imagenet pipeline using \tfdata, PyTorch DataLoader~\cite{pytorch-dataloader}, and NVIDIA DALI~\cite{nvidia-dali}. Table~\ref{tab:eval:resnet} shows the average time to process an epoch's worth of data with each framework running on a 64 core server (\texttt{n2-standard-64} on Google Cloud) with 256 GB of RAM, 500 GB local SSD, and NVIDIA Tesla T4 GPUs. The \tfdata pipeline is 1.9$\times$ faster than DataLoader, thanks to \tfdata's static and dynamic optimizations. For example, if we disable \mapdataset~+~\batchdataset fusion in \tfdata, performance drops to 448 seconds per epoch. Table~\ref{tab:eval:resnet} shows that \tfdata outperforms DALI on CPU or even with one GPU. When offloading computation to multiple GPUs, DALI achieves higher throughput, however using GPUs adds to the cost of input data processing and consumes GPU cores and memory that could otherwise be dedicated to model training.

In addition to comparing input pipeline throughput, it is useful to compare end-to-end model training time with different input data frameworks across heterogeneous platforms. The MLPerf Training competition provides the fairest comparison across ML systems as each submission is optimized by experts familiar with their performance knobs. For each benchmark, a cluster ranging from 8 accelerators to over 1000 accelerators was used to train the model to a target accuracy. Table~\ref{tab:mlperf_summary} summarizes the top MLPerf v0.7 training times achieved, categorized by the input pipeline framework used~\cite{mlperf-results-v7}. The end-to-end training times in Table~\ref{tab:mlperf_summary} do not provide an apples-to-apples performance comparison of input data frameworks, since the competition entries used different software frameworks (TensorFlow, PyTorch, MXNet) and hardware (TPUs, GPUs) to run model training computations. %
However, we can still draw two important takeaways from the end-to-end training times in Table~\ref{tab:mlperf_summary}. First, \tfdata is the only input processing framework that was used across all MLPerf benchmarks, including image and text workloads. This attests to \tfdata's flexibility. Other frameworks only achieved competitive results for a subset of benchmarks (e.g., DALI for image workloads and DataLoader for text workloads). Second, \tfdata is fast enough to avoid input bottlenecks across state-of-the-art models and hardware configurations, enabling training \resnet, \ssd, \transformer, and \bert in under 30 seconds. As shown in \S~\ref{sec:eval:large-scale}, the MLPerf workloads are not input-bound after applying \tfdata optimizations. In particular, the higher end-to-end training time with GNMT, is due to the TensorFlow model computation being slower  than the PyTorch implementation; the \tfdata part of the computation is not on the critical path.

\begin{table}[]
\small
\begin{tabularx}{\linewidth}{*1{>{\raggedright\arraybackslash}X}   *6{>{\centering\arraybackslash}X}}
  & \footnotesize{\resnet} &  \footnotesize{\ssd} & \footnotesize{\maskrcnn} & \footnotesize{\gnmt} & \footnotesize{\textsc{Trans-}\textsc{former}} & \footnotesize{\bert} \\
  \hline 
\tfdata     & 28.8 & 27.6 & 487.8 & 77.4 & 15.6 & 23.4 \\
DataLoader &  - & - & 627.6 & 42.6 & 37.2 & 48.6  \\
DALI & 49.8 & 49.2 & - & - & - & - \\
\end{tabularx}
\vspace{0.2cm}
\caption{Best MLPerf v0.7 competition training times (in seconds), categorized by the input data framework used. Entries with \tfdata, DataLoader, and DALI input pipelines use TensorFlow, PyTorch, and MXNet, resp., for model training.}
\label{tab:mlperf_summary}
\vspace{-0.5cm}
\end{table}

%% file: experience.tex
\section{Experience}
\label{sec:discussion}

At Google, we have been using \tfdata in training research and production ML models since 2017. As of today, the system implementation consists of over 15K lines of Python and over 40k lines of C++ (excluding test code). The \tfdata framework is used for data processing by the majority of TensorFlow training jobs in Google's fleet. These jobs run in production clusters, spanning a variety of application domains (e.g., image classification, translation, and video content recommendation) and using various types of ML training algorithms (e.g., supervised learning, reinforcement learning, and federated learning). \tfdata's generality has also facilitated novel research. For example, a creative approach to working around limited I/O bandwidth when training models and was implemented using three standard {\tfdata} transformations~\cite{data-echoing}. \tfdata was also used to automatically generate a data augmentation policy that achieved state-of-the-art results on image classification tasks~\cite{cubuk2019randaugment}.

To understand the characteristics of machine learning input data pipelines at scale, we studied millions of \tfdata jobs in Google's fleet over a one month period in 2020.
We show that input pipelines are highly diverse and frequently re-executed. We also identify several future research directions motivated by our findings, such as the opportunity to re-use input pipeline computation across jobs. %

\subsection{Fleet-wide Input Pipeline Analysis}\label{sec:experience:usage}

\textbf{Methodology:}
We instrument \tfdata to collect metrics such as the set of transformations applied in each job's input pipeline. For each job, we also record the bytes consumed and produced by each transformation in its input pipeline. $71\%$ of jobs define their input pipeline as a single \tfdata dataset, while the remaining jobs define their input processing logic across two or more \tfdata datasets. When an iterator is created for a \tfdata dataset, we fingerprint the dataset by computing a hash of its dataflow graph. We include the list of input file names in the hash calculation and exclude random seed values. We track the number of iterator creations for each unique hash over time. We also measure the total compute time for jobs and the compute time that jobs spend in \tfdata. The compute time is measured in normalized compute units and is the product of the time spent on a hardware resource -- such as a CPU or an accelerator core -- scaled by the compute capability of that resource. Our compute time metric is analogous to AWS's Elastic Compute Units (ECUs)~\cite{aws-faq-ecu}. We collect the metrics described above with full coverage for all \tfdata jobs, with one exception. Measuring the fraction of compute time spent in \tfdata requires a configuration flag to be set when jobs are launched. Due to configuration differences across jobs, we measured the fraction of compute time spent in \tfdata for 66\% of jobs, accounting for 75\% of total compute time across \tfdata jobs. For the remaining jobs, we assume that each job spends $10$\% of its total compute time in \tfdata, as this is the median time that jobs spend in the input pipeline (see Figure~\ref{fig:tfdata-gcu-fraction}).

Our analysis focuses on three key questions: 1) how frequently are various transformations used in an input pipeline, 2) how does the "shape" of data change as data flows through an input pipeline, and 3) how much computation is shared across input pipeline executions in our fleet?

\begin{figure}
\centering
\includegraphics[trim = 0.8cm 0.2cm 0cm 1cm, clip=true, width=\linewidth]{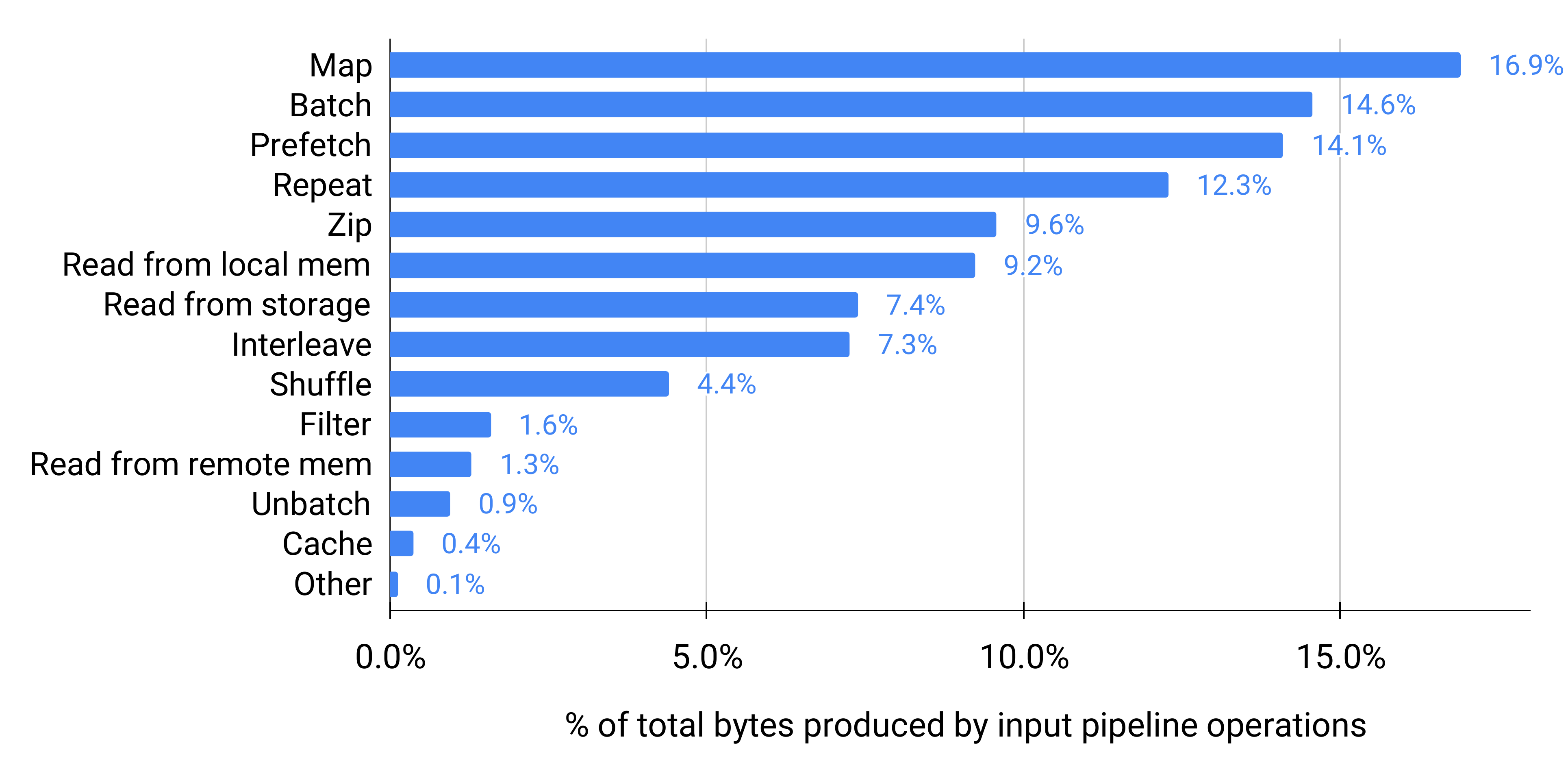}
\vspace{-0.6cm}
\caption{Types of input data pipeline operations and their prevalence, based on the bytes produced by each type of op.}
\vspace{-.1cm}
\label{fig:ops}
\end{figure}

\vspace{-.25cm}
\paragraph{Which datasets are most common?}{}
Figure~\ref{fig:ops} plots the relative frequency of \tfdata transformations across jobs, based on the number of bytes each transformation is applied on. The \mapdataset, \batchdataset, \prefetchdataset, \repeatdataset, and \zipdataset transformations are the five most commonly applied types of transformations, followed by reading input data from local memory and storage. We also study how many input pipelines rely on various \tfdata optimizations. On average, 77\% of input pipelines rely on parallel I/O optimizations by using the \interleavedataset transformation, 87\% of input pipelines rely on pipeline parallelism with the \prefetchdataset transformation, and 40\% of pipelines rely on parallelizing compute with the \mapdataset transformation (and its fusion with \batchdataset). Only 19\% of jobs use the \cachedataset transformation to cache input data in memory, though we later show that many more jobs could benefit from caching since many input pipelines are re-executed.

\vspace{-.25cm}
\paragraph{How does preprocessing affect data volume?}{
While some transformations, such as filtering, decrease the size of input data, machine learning jobs also commonly apply transformations that increase the size of data, such as decompressing and augmenting images to train image understanding models. To understand how the volume of data flowing through ML input pipelines varies with different transformations, we measure each input pipeline's ratio of bytes produced versus the bytes read from inputs sources. We compute this ratio for the end-to-end input pipeline of each job, as well as for each type of transformation applied in the job's input pipeline. When the bytes produced over bytes consumed ratio is less than one, it means that the input pipeline or transformation in this job decreases the data volume, whereas a ratio greater than one implies that the volume of data increases. %
}

\begin{figure}
\includegraphics[trim = 0.5cm 0cm 0.8cm 1cm,clip=true, width=\linewidth]
{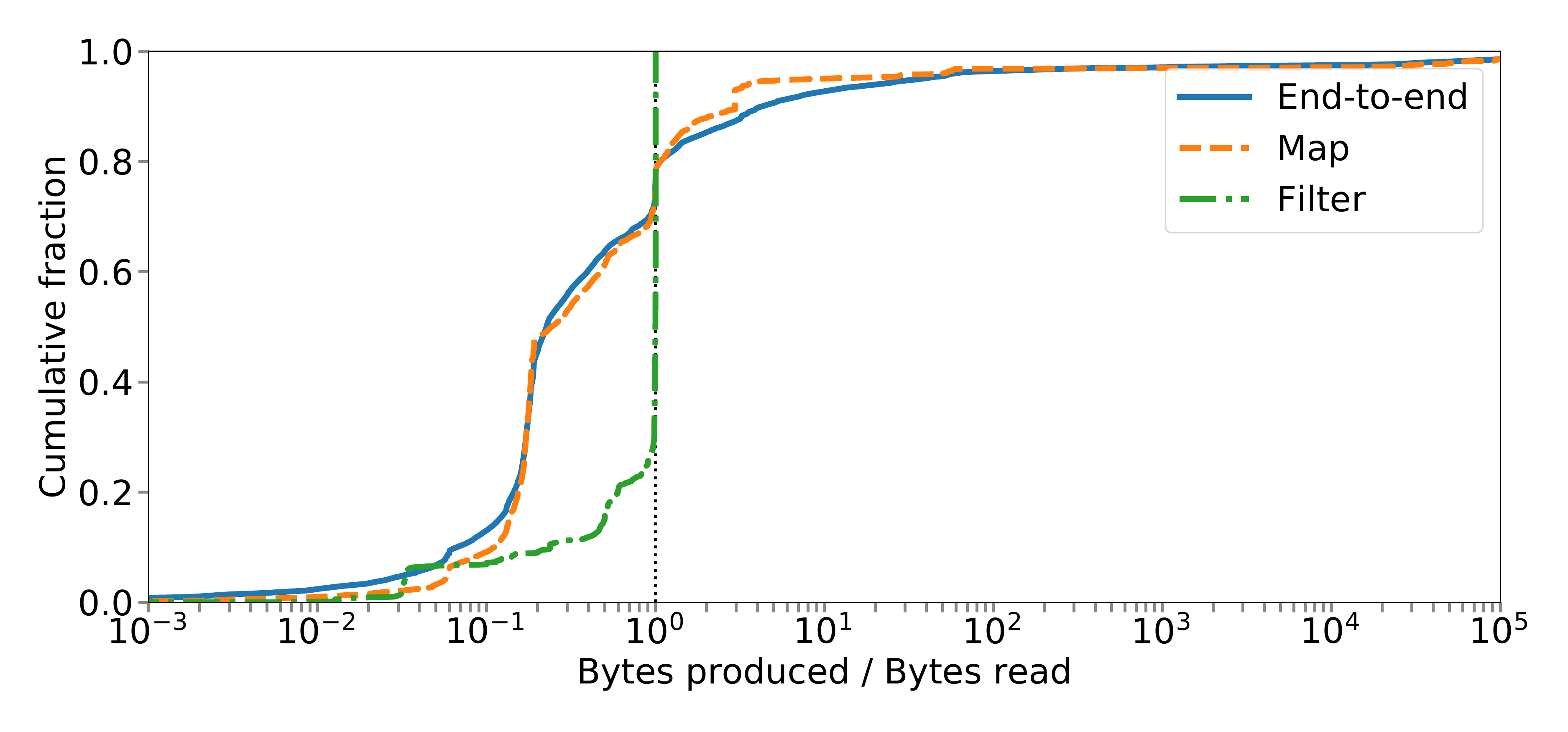}
\vspace{-0.8cm}
\caption{CDF showing how the ratio of bytes produced vs. bytes read varies for end-to-end input pipelines, \mapdataset, and \filterdataset transformations. For 75\% of jobs, preprocessing reduces the volume of data end-to-end.}
\vspace{-0.2cm}
\label{fig:data_volume}
\end{figure}

Figure~\ref{fig:data_volume} plots the CDF of the bytes produced over bytes consumed ratio across jobs for their end-to-end input pipeline, \mapdataset transformations, and \filterdataset transformations. For approximately 75\% of jobs, the volume of data produced by the input pipeline and fed to the model training stage is less than the volume of input data read. In other words, for most jobs, the materialized dataset used for training is smaller than the raw input data. For some jobs, decompressing and augmenting data results in high expansion of source data. Figure~\ref{fig:data_volume} shows that user-defined \mapdataset transformations, while preserving dataset cardinality, can decrease or expand data by over an order of magnitude for $13\%$ of jobs. \filterdataset transformations, which can modify dataset cardinality, discard more than $10\%$ of input data volume for approximately $23\%$ of jobs. For $8\%$ of jobs, more than half of the bytes fed into \filterdataset transformations are discarded. \filterdataset is also used to sanitize data, hence in $70\%$ of jobs, the transformation reduces the data by less than $1\%$.

\vspace{-.25cm}
\paragraph{How often are input pipelines re-executed?}
We observe that input pipelines are commonly re-executed and there is a sizeable opportunity to reuse input pipeline computation both within and across jobs. Some jobs rely on different permutations of the same dataset across iterators to improve convergence. To conservatively estimate the opportunity for computation reuse across input pipeline executions, we have excluded datasets that use the \shuffledataset transformation (57\% of \tfdata jobs) in this part our analysis.

\begin{figure}
\includegraphics[trim = 0.5cm 0.4cm 0cm 0cm,clip=true, width=\linewidth]
{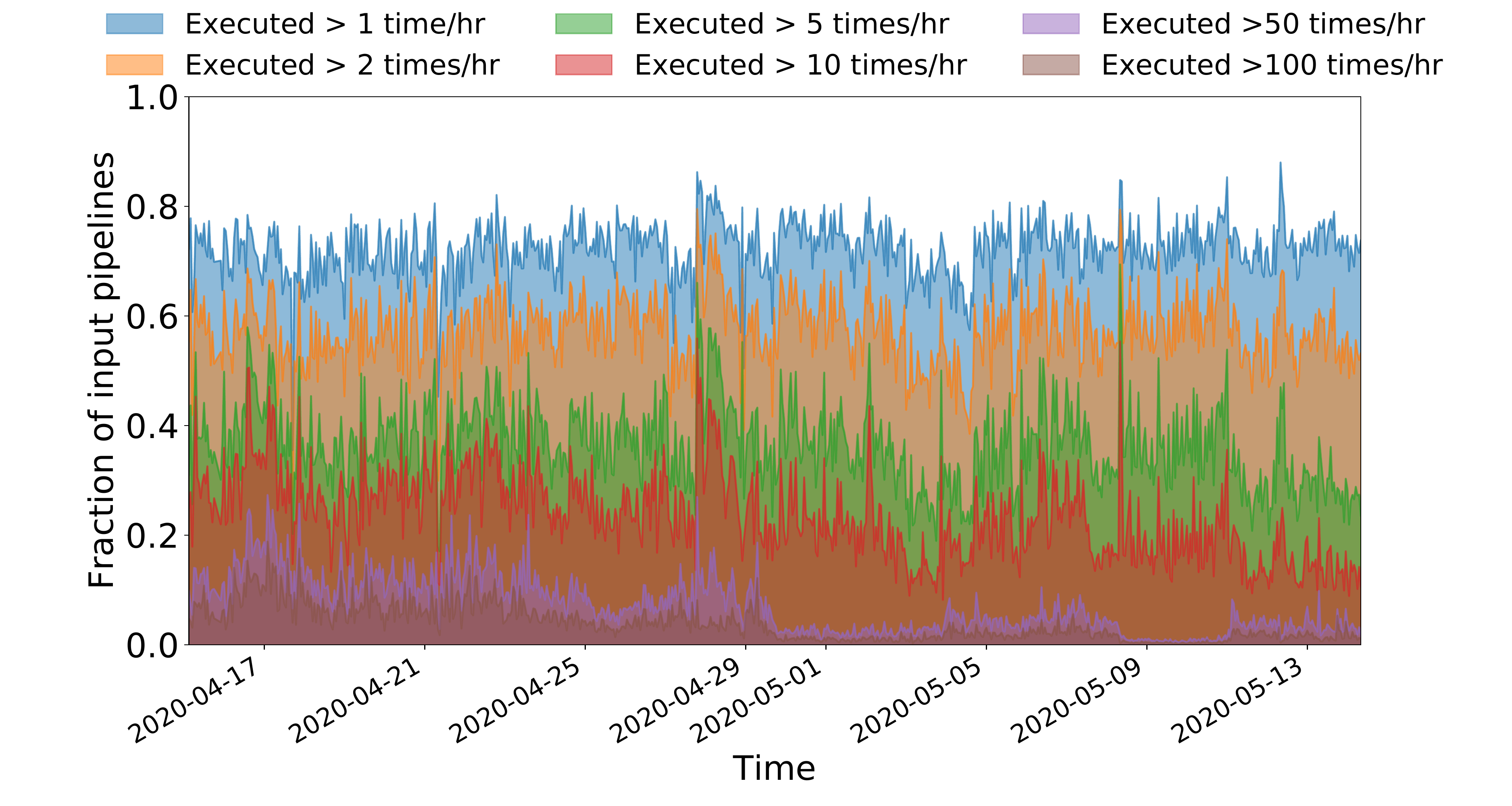}
\caption{Fraction of input pipelines executed more than $x$ times per hour, over time. Approx. 75\% of input pipelines are executed more than once in the same hour.}
\label{fig:iterator-duplicates}
\end{figure}

An input pipeline iteration begins by creating an iterator for a dataset definition. We record the number of iterator creations at the granularity of one hour time intervals for each dataset fingerprint (computed by hashing its dataflow graph).  
Figure~\ref{fig:iterator-duplicates} plots the fraction of input pipelines that are executed more than $x$ times in the same hour, over time. Approximately 75\% of input pipelines are executed more than once within the same hour and 5\% of input pipelines are executed more than 100 times within an hour. Re-execution of input pipelines can occur across epochs of a training job and also across jobs. For example, neural architecture search~\cite{nas} and hyper-parameter tuning both require training multiple models using the same input pipeline.

\begin{figure}
\includegraphics[trim = 0.5cm 0.3cm 0.9cm 0.6cm,clip=true, width=0.98\linewidth]
{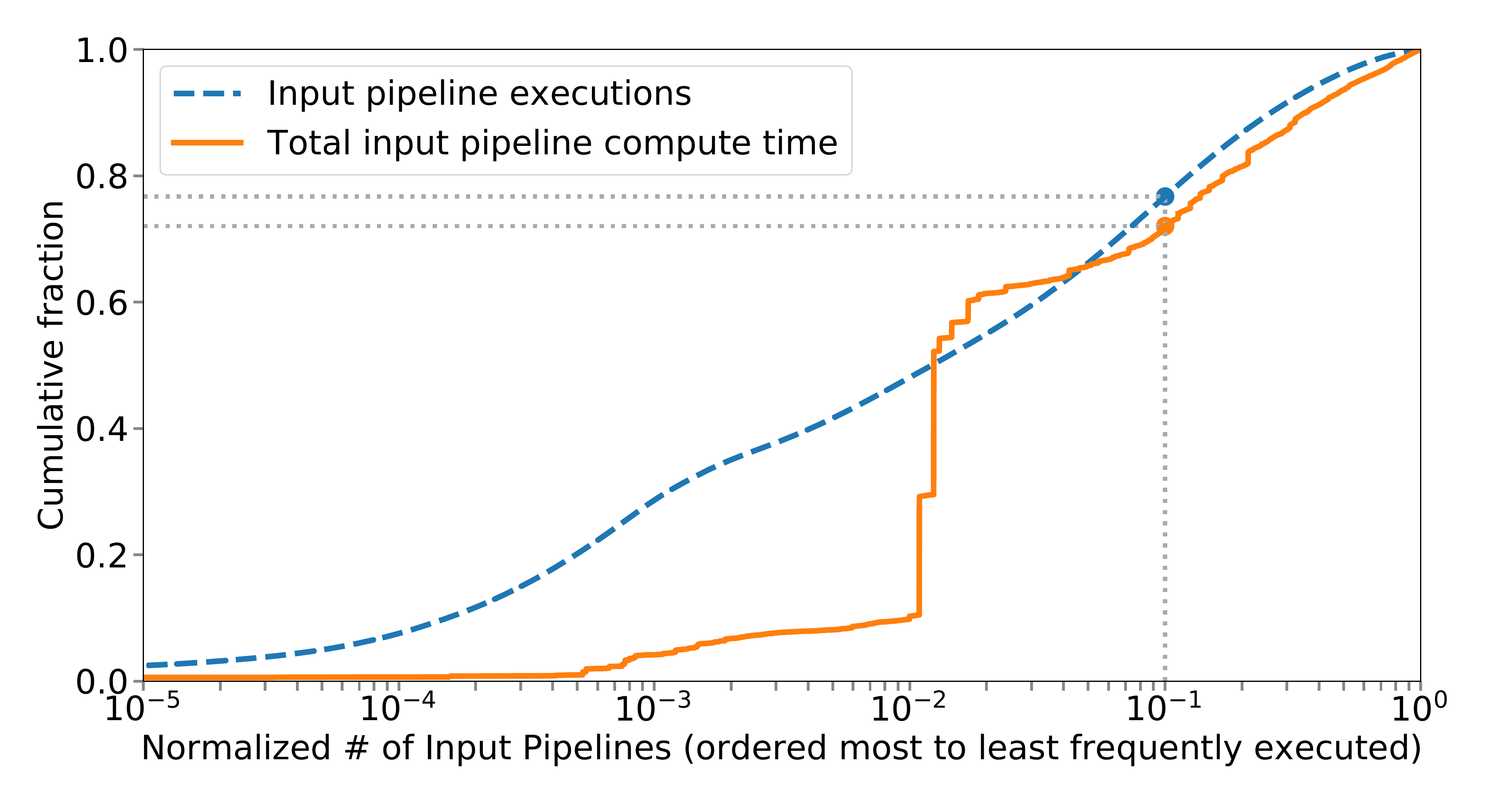}
\caption{CDF of input pipeline executions over a one month period. 10\% of pipelines account for 77\% of total input pipeline executions and 72\% of compute resources.} 
\label{fig:fingerprint-gcus}
\end{figure}

Having found that many input pipelines are re-executed, we next quantify the opportunity for reusing input pipeline computation by caching materialized datasets. Figure~\ref{fig:fingerprint-gcus} plots the cumulative distribution of input pipeline executions over the one month time span of our study, with input pipelines ordered from most to least frequently executed. We also show the CDF of the compute resources spent executing these pipelines. %
Figure~\ref{fig:fingerprint-gcus} shows that by caching the top $10\%$ of materialized datasets, we can capture 72\% of CPU resources used for computing \tfdata datasets across all jobs that executed in the one month period. The steepness of the CDF curves indicates that some datasets are particularly frequently executed and consumed significant resources. Only 10\% of input pipelines are re-executed across multiple jobs. 1\% of input pipelines are executed by more than 25 different jobs and the largest cross-job sharing we observed was approximately 50,000 jobs executing the same input pipeline. %
However, our analysis conservatively estimates the opportunity for reuse since it only counts re-executions of pipelines with identical end-to-end transformation graphs. We anticipate further opportunities to reuse computation across jobs by considering input pipeline sub-graphs.

\subsection{Implications for Future Research}
\label{ss:discussion:future}

\paragraph{Datasets as a service} 
We showed that input pipelines are frequently re-executed, yet only 19\% of jobs in our analysis used the \cachedataset transformation. It is often challenging for users to decide if and where to apply caching as there are several factors to consider: the cost-benefit of caching the data -- spending RAM to save CPU and possibly improve throughput -- and the impact of caching on training quality -- in general, results of randomized transformation (such as \shuffledataset) should not be cached. Navigating the compute-storage trade-off and estimating the benefit of caching on end-to-end performance and downstream accelerator utilization for ML training jobs is a complex task for users~\cite{nectar}. Hence, automating \cachedataset insertion in input pipelines is important~\cite{helix}. %
Furthermore, since input pipelines can be shared across jobs, designing a dataset caching service to re-use input pipeline computations across jobs is a promising future direction. Quiver~\cite{quiver} and CoorDL~\cite{coordl} already optimize source dataset caching for ML training. Several systems have shown that caching data across jobs greatly improves performance for big data analytics~\cite{nectar, pacman, ec-cache, quartet, tachyon}. 

\vspace{-.25cm}
\paragraph{Processing data closer to storage}
Figure~\ref{fig:data_volume} showed that data preprocessing reduces the volume of data for 75\% of jobs. For $14\%$ of jobs, the volume of data fed into the model for training is less than 10\% of bytes read from storage. As input data for ML jobs commonly resides in remote storage, such as a distributed file system or cloud object store, this means that more data than necessary is sent over the network during ML training. %
Designing ML data processing systems that apply projections closer to storage is a promising way to reduce data transfers. Using columnar data formats is another well-known approach to enable reading only the relevant fields in a record~\cite{parquet}. We are exploring this approach to improve data ingestion efficiency for ML jobs. %

\vspace{-.25cm}
\paragraph{Addressing host bottlenecks} 
Some input pipelines require significant CPU and memory resources to produce their data. When a host machine isn't powerful enough to generate input data at the rate the attached accelerator(s) consume the data, the accelerator(s) idle and slow down model training. To solve this problem, we are currently exploring the disaggregation of data processing from model training, by enabling users feed accelerators from input workers distributed across multiple hosts. The number of input workers can scale up or down as needed to keep up with the accelerators, independent of the number of accelerators attached to one host. Another approach to address host resource bottlenecks for input data processing is to offload data preprocessing computations to accelerators~\cite{nvidia-dali}. %

\vspace{-.25cm}
\paragraph{Data processing for online inference}
\tfdata targets the input processing needs of ML training jobs. However,  input pipeline efficiency is also critical for ML inference. Online inference pipelines perform fewer model computations per input element since only a forward pass of the model is required, whereas training also requires backpropagation. Hence, although not all input pipeline transformations applied during training -- such as data augmentations -- are applied when serving a model, the input pipeline for inference still presents a significant fraction of total work. Inference jobs need a different input pipeline design as the primary performance objective is to optimize the latency of individual requests rather than overall throughput. This implies less buffering, no shuffling, and a different approach to batching to balance request latency with accelerator efficiency.

%% file: related.tex
\section{Related Work}
\label{sec:relatedwork}

Kakarapathy \textit{et al.} made the case for building a single, unified system for data loading that could be shared between multiple machine learning jobs, and potentially between different frameworks as well~\cite{unifying-data-loading-hotcloud19}. Their observation that much I/O and preprocessing work can be shared between jobs agrees with our findings in \S~\ref{ss:discussion:future}. %
By contrast, our work on {\tfdata} has focused on a more general programming model, to enable users to build different preprocessing schemes.

Our inspiration for {\tfdata}'s programming model drew from the successful application of LINQ~\cite{linq} to parallel processing with PLINQ~\cite{plinq}, big-data cluster processing with DryadLINQ~\cite{dryadlinq}, and stream processing with Naiad~\cite{murray2013naiad}. Many transformations in {\tfdata} have direct equivalents in LINQ, though we added order-sensitive transformations (e.g., \batchdataset, \shuffledataset, and \interleavedataset) to support ML training algorithms. %
Optimus~\cite{optimus}, which added dynamic graph rewriting support to DryadLINQ, is similar to the automatic optimization framework that we described in \S~\ref{ss:design:auto-opt}. Optimus focused on reducing network I/O in distributed big-data queries, whereas the bottleneck in {\tfdata} applications tends to be the host CPU, and our optimizations aim to reduce the wall-clock execution time of code within a single machine. Dandelion extended LINQ with the ability to run on accelerator devices such as GPUs and FPGAs~\cite{dandelion}, using the PTask abstraction to manage the accelerators~\cite{ptask}. Respectively, Dandelion and PTask provide a simple programming model and optimized implementation that hides data movement between the host and accelerator devices, similar to how {\tfdata} uses {\prefetchdataset} to mask copies. Dandelion goes further than {\tfdata} in using functional transformations to represent all computation -- not just the input pipeline -- while {\tfdata} interoperates with existing ML frameworks such as TensorFlow~\cite{tensorflow}, Pytorch~\cite{pytorch}, and JAX~\cite{jax} by using their existing programming models for the training loop. %

The design, implementation, and optimization of {\tfdata} all bear similarities to how SQL is used in a relational database management system (RDBMS). A related strand of work has investigated how to push machine learning computations into SQL, and optimize across the boundary between relational data and linear algebra. The MADlib analytics library pushes various learning algorithms into an existing RDBMS~\cite{madlib}. MADlib uses existing SQL constructs for orchestration -- i.e.\ defining both the input pipeline and the ``driver program'' (or training loop) -- and provides a C++ abstraction layer for plugging in user-defined functions that call high-performance numerical libraries. %
By building {\tfdata} into TensorFlow and using its \texttt{Tensor} type to represent values, we achieved efficient interoperability for free. More recently, Karanasos \textit{et al.} introduced Raven, which integrates the ONNX Runtime for machine learning into Microsoft SQL Server~\cite{karanasos2019extending}. Raven focuses on ML inference for SQL-based analytic pipelines, achieving better performance by pushing linear algebra operators into earlier stages of the query plan and using ONNX Runtime to offload computation to accelerators. The model-related optimizations in {\tfdata} are more conservative than Raven's, because the model is mutable at training time, but the ideas in Raven would be useful for applications like knowledge distillation~\cite{distillation}, where inference on one model generates features for training another model.

Several related projects have investigated the problem of automatically tuning dataflow workloads. SEDA addresses the problem of dynamic resource allocation to stages, using a simple scheme that adds threads to a stage when its queue length exceeds a threshold, and removes them when they idle for a period~\cite{seda}. By contrast, {\tfdata} tunes the performance of each stage based on the predicted effect on end-to-end performance. %
The DS2 scaling controller for dataflow-based stream processing attempts to find the minimum parallelism for each stage in a dataflow graph that will enable it to consume data at the rates of all the sources~\cite{ds2}. Like DS2, {\tfdata} uses lightweight instrumentation of ``useful'' processing time in each transformation to make scaling decisions, but we additionally model memory consumption as a possible bottleneck resource to avoid excessive buffering.

%% file: conclusion.tex
\section{Conclusion}
\label{sec:conclusion}
 
We presented \tfdata, a framework for building and executing efficient input data processing pipelines for machine learning jobs at scale. \tfdata's programming model enables users to build diverse input pipelines by composing and customizing operators. \tfdata executes input pipelines as dataflow graphs and applies static optimizations that improve end-to-end training time for state-of-the-art models. For example, input pipeline parallelism and software pipelining improve \resnet training time by over $10\times$ and other \tfdata optimizations such as operator fusion provide an additional $2\times$ improvement. %
We developed an analytical approach to automatically tune internal buffer sizes and the degree of parallelism in input pipelines. These dynamic optimizations achieve comparable performance to expert-tuned input pipelines while relieving users from the burden of manually tuning parameters. %

Our fleet-wide analysis of \tfdata usage across millions of real jobs at Google quantified several aspects of ML data processing at scale, namely its resource footprint, diversity, and extent of redundant computation. Our findings motivate future work on sharing computation across jobs and pushing data projection to the storage layer. %

%% file: acknowledgements.tex
\section*{Acknowledgements}

We thank Paul Barham, Chandu Thekkath, Vijay Vasudevan, Martin Abadi, Sudip Roy, Dehao Chen, and our anonymous reviewers for their helpful feedback on this work. We gratefully acknowledge Andrew Audibert, Brennan Saeta, Fei Hu, Piotr Padlewski, Rachel Lim, Rohan Jain, Saurabh Saxena, and Shivani Agrawal for their engineering contributions to \tfdata.

%% file: main.bib
@InProceedings{murray2013naiad,
author = {Murray, Derek G. and McSherry, Frank and Isaacs, Rebecca and Isard, Michael and Barham, Paul and Abadi, Mart\'{i}n},
title = {Naiad: A Timely Dataflow System},
booktitle = {Proceedings of the 24th ACM Symposium on Operating Systems Principles (SOSP)},
year = {2013},
month = {November},
publisher = {ACM},
%url = {https://www.microsoft.com/en-us/research/publication/naiad-a-timely-dataflow-system-2/},
edition = {Proceedings of the 24th ACM Symposium on Operating Systems Principles (SOSP)},
}

@inproceedings{imagenet,
        AUTHOR = {Deng, Jia and Dong, W. and Socher, R. and Li, L.-J. and Li, K. and Fei-Fei, L.},
        TITLE = {{ImageNet: A Large-Scale Hierarchical Image Database}},
        BOOKTITLE = {Proceedings of CVPR},
        YEAR = {2009},
        %BIBSOURCE = "http://www.image-net.org/papers/imagenet_cvpr09.bib"
}

@inproceedings{best-practices-cnns,
author = {Simard, Patrice Y. and Steinkraus, Dave and Platt, John C.},
title = {Best Practices for Convolutional Neural Networks Applied to Visual Document Analysis},
year = {2003},
%isbn = {0769519601},
publisher = {IEEE Computer Society},
%address = {USA},
booktitle = {Proceedings of ICDAR},
pages = {958},
numpages = {1},
%series = {ICDAR ’03}
}

@inproceedings{dryadlinq,
author = {Yu, Yuan and Isard, Michael and Fetterly, Dennis and Budiu, Mihai and Erlingsson, \'{U}lfar and Gunda, Pradeep Kumar and Currey, Jon},
title = {DryadLINQ: A System for General-Purpose Distributed Data-Parallel Computing Using a High-Level Language},
year = {2008},
booktitle = {Proceedings of OSDI},
pages = {1--14},
%numpages = {14},
%location = {San Diego, California},
%series = {OSDI’08}
}

@inproceedings{unifying-data-loading-hotcloud19,
author = {Aarati Kakaraparthy and Abhay Venkatesh and Amar Phanishayee and Shivaram Venkataraman},
title = {The Case for Unifying Data Loading in Machine Learning Clusters},
booktitle = {Proceedings of HotCloud},
year = {2019},
address = {Renton, WA},
%url = {https://www.usenix.org/conference/hotcloud19/presentation/kakaraparthy},
%publisher = {{USENIX} Association},
%month = jul,
}

@article{helix,
author = {Xin, Doris and Ma, Litian and Liu, Jialin and Macke, Stephen and Song, Shuchen and Parameswaran, Aditya},
title = {Helix: Accelerating Human-in-the-Loop Machine Learning},
year = {2018},
issue_date = {August 2018},
publisher = {VLDB Endowment},
volume = {11},
number = {12},
%issn = {2150-8097},
%url = {https://doi.org/10.14778/3229863.3236234},
%doi = {10.14778/3229863.3236234},
journal = {Proc. VLDB Endow.},
month = aug,
pages = {1958--1961},
numpages = {4}
}

@article{tpuv2v3-cacm,
author = {Jouppi, Norman P. and Yoon, Doe Hyun and Kurian, George and Li, Sheng and Patil, Nishant and Laudon, James and Young, Cliff and Patterson, David},
title = {A Domain-Specific Supercomputer for Training Deep Neural Networks},
year = {2020},
issue_date = {July 2020},
publisher = {Association for Computing Machinery},
volume = {63},
number = {7},
journal = {Commun. ACM},
month = jun,
pages = {67-78},
}

@inproceedings{spark-streaming,
author = {Zaharia, Matei and Das, Tathagata and Li, Haoyuan and Hunter, Timothy and Shenker, Scott and Stoica, Ion},
title = {Discretized Streams: Fault-Tolerant Streaming Computation at Scale},
year = {2013},
%booktitle = {Proceedings of the Twenty-Fourth ACM Symposium on Operating Systems Principles},
booktitle = {Proceedings of SOSP},
pages = {423-438},
numpages = {16},
%series = {SOSP '13}
}

@misc{coordl,
      title={Analyzing and Mitigating Data Stalls in DNN Training}, 
      author={Jayashree Mohan and Amar Phanishayee and Ashish Raniwala and Vijay Chidambaram},
      year={2021},
      eprint={2007.06775},
      archivePrefix={arXiv},
      primaryClass={cs.DC}
}

@inproceedings{data-validation-mlsys,
  author    = {Eric Breck and
               Neoklis Polyzotis and
               Sudip Roy and
               Steven Whang and
               Martin Zinkevich},
  title     = {Data Validation for Machine Learning},
  booktitle = {Proceedings of Machine Learning and Systems (MLSys) 2019},
  year      = {2019},
}

@misc{aws-faq-ecu,
   author={Amazon},
   title={{Amazon EC2 FAQs}},
   howpublished={\url{https://aws.amazon.com/ec2/faqs}},
   year=2020
}

@misc{avro,
   author={{Apache Software Foundation}},
   title={{Avro}},
   howpublished={\url{https://avro.apache.org/docs/1.2.0}},
   year=2012
}

@misc{parquet,
   author={{Apache Software Foundation}},
   title={{Parquet}},
   howpublished={\url{https://parquet.apache.org/}},
   year=2018
}

@misc{beam,
   title={{Apache Beam: An advanced unified programming model}},
   howpublished={\url{https://beam.apache.org/}},
   year=2020
}

@misc{flume,
   title={{Apache Flume}},
   howpublished={\url{https://flume.apache.org/}},
   year=2020
}

@misc{data-echoing,
    title={{Faster Neural Network Training with Data Echoing}},
    author={Dami Choi and Alexandre Passos and Christopher J. Shallue and George E. Dahl},
    year={2019},
    eprint={1907.05550},
    archivePrefix={arXiv},
    primaryClass={cs.LG}
}

@misc{nvidia-dali,
   author={Joaquin Anton Guirao and Krzysztof Łęcki and Janusz Lisiecki and Serge Panev and Michał Szołucha and Albert Wolant and Michał Zientkiewicz},
   title={{Fast AI Data Preprocessing with NVIDIA DALI}},
   howpublished={\url{https://devblogs.nvidia.com/fast-ai-data-preprocessing-with-nvidia-dali}},
   year=2019
}

@misc{pytorch-dataloader,
   author={Torch Contributors},
   title={{PyTorch Docs: torch.utils.data}},
   howpublished={\url{https://pytorch.org/docs/stable/data.html}},
   year=2019
}

@misc{tfrecord,
   author={{TensorFlow}},
   title={{TFRecord and tf.Example}},
   howpublished={\url{https://www.tensorflow.org/tutorials/load_data/tfrecord}},
   year=2020
}

@misc{google-cloud-pricing,
   author={{Google}},
   title={{Google Cloud: All Pricing}},
   howpublished={\url{https://cloud.google.com/compute/all-pricing}},
   year=2020
}

@misc{aws-pricing,
   author={{Amazon}},
   title={{Amazon EC2 Pricing}},
   howpublished={\url{https://aws.amazon.com/ec2/pricing/}},
   year=2020
}

@inproceedings{linq,
author = {Meijer, Erik and Beckman, Brian and Bierman, Gavin},
title = {{LINQ}: Reconciling Object, Relations and {XML} in the {.NET Framework}},
year = {2006},
%isbn = {1595934340},
%publisher = {Association for Computing Machinery},
%address = {New York, NY, USA},
%url = {https://doi.org/10.1145/1142473.1142552},
%doi = {10.1145/1142473.1142552},
booktitle = {Proceedings of SIGMOD},
pages = {706},
numpages = {1},
%location = {Chicago, IL, USA},
%series = {SIGMOD ’06}
}

@misc{java-streams,
   author={{Java}},
   title={{Stream API}},
   howpublished={\url{https://docs.oracle.com/javase/8/docs/api/java/util/stream/package-summary.html}},
   year=2020
}

@inproceedings{mapreduce,
title	= {MapReduce: Simplified Data Processing on Large Clusters},
author	= {Jeffrey Dean and Sanjay Ghemawat},
year	= {2004},
booktitle	= {Proceedings of OSDI},
pages	= {137--150},
}

@inproceedings{spark,
author = {Zaharia, Matei and Chowdhury, Mosharaf and Franklin, Michael J. and Shenker, Scott and Stoica, Ion},
title = {Spark: Cluster Computing with Working Sets},
year = {2010},
%publisher = {USENIX Association},
%address = {USA},
booktitle = {Proceedings of HotCloud},
%pages = {10},
%numpages = {1},
%location = {Boston, MA},
%series = {HotCloud’10}
}

@article{sgd,
author = "Robbins, Herbert and Monro, Sutton",
%doi = "10.1214/aoms/1177729586",
fjournal = "Annals of Mathematical Statistics",
journal = "Ann. Math. Statist.",
month = "09",
number = "3",
pages = "400--407",
publisher = "The Institute of Mathematical Statistics",
title = "A Stochastic Approximation Method",
%url = "https://doi.org/10.1214/aoms/1177729586",
volume = "22",
year = "1951"
}

@article{volcano,
author = {Graefe, Goetz},
title = {Volcano: An Extensible and Parallel Query Evaluation System},
year = {1994},
%issue_date = {February 1994},
%address = {USA},
volume = {6},
number = {1},
%issn = {1041-4347},
%url = {https://doi.org/10.1109/69.273032},
%doi = {10.1109/69.273032},
journal = {IEEE Trans. on Knowledge and Data Engineering},
month = {Feb},
pages = {120--135},
%numpages = {16},
}

@InProceedings{bottou-curiously-fast-convergence,
  author = {Leon Bottou},
  title = {{Curiously Fast Convergence of some Stochastic Gradient Descent Algorithms}},
  booktitle = {Proceedings of the Symposium on Learning and Data Science},
  year = {2009}
}

@inproceedings{autograph,
title	= {AutoGraph: Imperative-style Coding with Graph-based Performance},
author	= {Dan Moldovan and James Decker and Fei Wang and Andrew Johnson and Brian Lee and Zack Nado and D Sculley and Tiark Rompf and Alexander B Wiltschko},
year	= {2019},
booktitle	= {SysML}
}

@article{mlir,
  author    = {Chris Lattner and
               Jacques A. Pienaar and
               Mehdi Amini and
               Uday Bondhugula and
               River Riddle and
               Albert Cohen and
               Tatiana Shpeisman and
               Andy Davis and
               Nicolas Vasilache and
               Oleksandr Zinenko},
  title     = {{MLIR:} {A} Compiler Infrastructure for the End of Moore's Law},
  journal   = {CoRR},
  year      = {2020},
  url       = {https://arxiv.org/abs/2002.11054},
  archivePrefix = {arXiv},
 % eprint    = {2002.11054},
 % timestamp = {Tue, 03 Mar 2020 14:32:13 +0100},
 % biburl    = {https://dblp.org/rec/journals/corr/abs-2002-11054.bib},
 % bibsource = {dblp computer science bibliography, https://dblp.org}
}

@misc{grappler,
   author={{TensorFlow}},
   title={{TensorFlow Graph Optimizations}},
   howpublished={\url{https://research.google/pubs/pub48051.pdf}},
   year=2019
}

@inproceedings{deforestation,
author = {Wadler, Philip},
title = {Deforestation: Transforming Programs to Eliminate Trees},
year = {1988},
publisher = {North-Holland Publishing Co.},
address = {NLD},
booktitle = {Proceedings of the Second European Symposium on Programming},
pages = {231--248},
numpages = {18},
%location = {Nancy, France}
}

@InProceedings{pfor,
  title = 	 {Static Automatic Batching In {T}ensor{F}low},
  author = 	 {Agarwal, Ashish},
  booktitle = 	 {Proceedings of ICML},
  pages = 	 {92--101},
  year = 	 {2019},
}

@article{mm1k,
  author =       {John E. Shore},
  title =        {The lazy repairman and other models: {Performance}
                 collapse due to overhead in simple, single-server
                 queuing systems},
  volume =       {9},
  number =       {2},
  pages =        {217--224},
  year =         {1980},
  journal =     {ACM SIGMETRICS Performance Evaluation Review},
}

@inproceedings{resnet,
  author    = {Kaiming He and
               Xiangyu Zhang and
               Shaoqing Ren and
               Jian Sun},
  title     = {Deep Residual Learning for Image Recognition},
  booktitle = {Proceedings of CVPR},
  pages     = {770--778},
  publisher = {{IEEE} Computer Society},
  year      = {2016},
  %url       = {https://doi.org/10.1109/CVPR.2016.90},
  %doi       = {10.1109/CVPR.2016.90},
  %timestamp = {Wed, 16 Oct 2019 14:14:50 +0200},
  %biburl    = {https://dblp.org/rec/conf/cvpr/HeZRS16.bib},
  %bibsource = {dblp computer science bibliography, https://dblp.org}
}

@inproceedings{coco,
title = {Microsoft COCO: Common Objects in Context},
author = {Tsung-Yi Lin and Michael Maire and Serge Belongie and James Hays and Pietro Perona and Deva Ramanan and Piotr Dollár and C. Lawrence Zitnick},
%url = {/se3/wp-content/uploads/2014/09/coco_eccv.pdf, http://mscoco.org},
year = {2014},
%date = {2014-01-01},
booktitle = {Proceedings of ECCV},
%address = {Zürich},
%note = {Oral},
%keywords = {}
}

@inproceedings{nas,
  author    = {Barret Zoph and
               Quoc V. Le},
  title     = {Neural Architecture Search with Reinforcement Learning},
  booktitle = {Proceedings of ICLR},
  year = {2017},
}

@inproceedings{ssd,
  title={{SSD: Single shot multibox detector}},
  author={Liu, Wei and Anguelov, Dragomir and Erhan, Dumitru and Szegedy, Christian and Reed, Scott and Fu, Cheng-Yang and Berg, Alexander C},
  booktitle={Proceedings of ECCV},
  pages={21--37},
  year={2016},
  organization={Springer}
}

@inproceedings{nectar,
author = {Gunda, Pradeep Kumar and Ravindranath, Lenin and Thekkath, Chandu and Yu, Yuan and Zhuang, Li},
title = {Nectar: Automatic Management of Data and Computation in Datacenters},
booktitle = {Proceedings of OSDI},
year = {2010},
}

@inproceedings {quiver,
author = {Abhishek Vijaya Kumar and Muthian Sivathanu},
title = {Quiver: An Informed Storage Cache for Deep Learning},
booktitle = {Proceedings of FAST},
year = {2020},
pages = {283--296},
}

@inproceedings {ec-cache,
author = {K. V. Rashmi and Mosharaf Chowdhury and Jack Kosaian and Ion Stoica and Kannan Ramchandran},
title = {EC-Cache: Load-Balanced, Low-Latency Cluster Caching with Online Erasure Coding},
booktitle = {Proceedings of OSDI},
year = {2016},
pages = {401--417},
}

@inproceedings{pacman,
author = {Ananthanarayanan, Ganesh and Ghodsi, Ali and Wang, Andrew and Borthakur, Dhruba and Kandula, Srikanth and Shenker, Scott and Stoica, Ion},
title = {PACMan: Coordinated Memory Caching for Parallel Jobs},
year = {2012},
booktitle = {Proceedings of NSDI},
pages = {20},
}

@inproceedings {quartet,
author = {Francis Deslauriers and Peter McCormick and George Amvrosiadis and Ashvin Goel and Angela Demke Brown},
title = {Quartet: Harmonizing Task Scheduling and Caching for Cluster Computing},
booktitle = {Proceedings of HotStorage},
year = {2016},
}

@inproceedings{tachyon,
author = {Li, Haoyuan and Ghodsi, Ali and Zaharia, Matei and Shenker, Scott and Stoica, Ion},
title = {Tachyon: Reliable, Memory Speed Storage for Cluster Computing Frameworks},
year = {2014},
booktitle = {Proceedings of SoCC},
pages = {1--15},
}

@article{maskrcnn,
  author    = {Kaiming He and
               Georgia Gkioxari and
               Piotr Doll{\'{a}}r and
               Ross B. Girshick},
  title     = {Mask {R-CNN}},
  journal   = {CoRR},
  year      = {2017},
  url       = {http://arxiv.org/abs/1703.06870},
  archivePrefix = {arXiv},
  %eprint    = {1703.06870},
  %timestamp = {Mon, 13 Aug 2018 16:46:36 +0200},
  %biburl    = {https://dblp.org/rec/journals/corr/HeGDG17.bib},
  %bibsource = {dblp computer science bibliography, https://dblp.org}
}

@inproceedings{transformer,
title = {{Attention is All you Need}},
author = {Vaswani, Ashish and Shazeer, Noam and Parmar, Niki and Uszkoreit, Jakob and Jones, Llion and Gomez, Aidan N and Kaiser, \L ukasz and Polosukhin, Illia},
booktitle = {Advances in Neural Information Processing Systems 30},
pages = {5998--6008},
year = {2017},
}

@inproceedings{bert,
    title = "{BERT}: Pre-training of Deep Bidirectional Transformers for Language Understanding",
    author = "Devlin, Jacob  and
      Chang, Ming-Wei  and
      Lee, Kenton  and
      Toutanova, Kristina",
    booktitle = "Proceedings of the 2019 Conference of the North {A}merican Chapter of the Association for Computational Linguistics: Human Language Technologies, Volume 1",
    month = jun,
    year = "2019",
    %address = "Minneapolis, Minnesota",
    %publisher = "Association for Computational Linguistics",
    %url = "https://www.aclweb.org/anthology/N19-1423",
    pages = "4171--4186",
}

@article{mlperf,
  title={{MLPerf} training benchmark},
  author={Mattson, Peter and Cheng, Christine and Coleman, Cody and Diamos, Greg and Micikevicius, Paulius and Patterson, David and Tang, Hanlin and Wei, Gu-Yeon and Bailis, Peter and Bittorf, Victor and others},
  journal={arXiv preprint arXiv:1910.01500},
  year={2019}
}

@inproceedings{optimus,
  author    = {Qifa Ke and
               Michael Isard and
               Yuan Yu},
  editor    = {Zdenek Hanz{\'{a}}lek and
               Hermann H{\"{a}}rtig and
               Miguel Castro and
               M. Frans Kaashoek},
  title     = {Optimus: a dynamic rewriting framework for data-parallel execution
               plans},
  booktitle = {Proceedings of EuroSys},
  pages     = {15--28},
  %publisher = {{ACM}},
  year      = {2013},
  %url       = {https://doi.org/10.1145/2465351.2465354},
  %doi       = {10.1145/2465351.2465354},
  %timestamp = {Tue, 06 Nov 2018 16:58:31 +0100},
  %biburl    = {https://dblp.org/rec/conf/eurosys/KeIY13.bib},
  %bibsource = {dblp computer science bibliography, https://dblp.org}
}

@inproceedings{ds2,
author = {Kalavri, Vasiliki and Liagouris, John and Hoffmann, Moritz and Dimitrova, Desislava and Forshaw, Matthew and Roscoe, Timothy},
title = {Three Steps is All You Need: Fast, Accurate, Automatic Scaling Decisions for Distributed Streaming Dataflows},
year = {2018},
%isbn = {9781931971478},
%publisher = {USENIX Association},
%address = {USA},
booktitle = {Proceedings of OSDI},
pages = {783--798},
numpages = {16},
%location = {Carlsbad, CA, USA},
%series = {OSDI’18}
}

@misc{wmt16,
   author={{WMT}},
   title={{1st Conference on Machine Translation}},
   howpublished={\url{http://statmt.org/wmt16}},
   year=2016
}

@misc{wmt17,
   author={{WMT}},
   title={{2nd Conference on Machine Translation}},
   howpublished={\url{http://statmt.org/wmt17}},
   year=2017
}

@article{madlib,
author = {Hellerstein, Joseph M. and R\'{e}, Christoper and Schoppmann, Florian and Wang, Daisy Zhe and Fratkin, Eugene and Gorajek, Aleksander and Ng, Kee Siong and Welton, Caleb and Feng, Xixuan and Li, Kun and Kumar, Arun},
title = {The MADlib Analytics Library: Or MAD Skills, the SQL},
year = {2012},
issue_date = {August 2012},
publisher = {VLDB Endowment},
volume = {5},
number = {12},
%issn = {2150-8097},
%url = {https://doi.org/10.14778/2367502.2367510},
%doi = {10.14778/2367502.2367510},
journal = {Proc. VLDB Endow.},
month = aug,
pages = {1700--1711},
numpages = {12}
}

@misc{protobuf,
   %author={{Google}},
   title={{Protocol Buffers}},
   howpublished={\url{https://developers.google.com/protocol-buffers}}
}

@article{gnmt,
  title={{Google's neural machine translation system: Bridging the gap between human and machine translation}},
  author={Wu, Yonghui and Schuster, Mike and Chen, Zhifeng and Le, Quoc V and Norouzi, Mohammad and Macherey, Wolfgang and Krikun, Maxim and Cao, Yuan and Gao, Qin and Macherey, Klaus and others},
  journal={arXiv preprint arXiv:1609.08144},
  year={2016}
}

@misc{mxnet_dataio,
   author={{MXNET}},
   title={{Designing Efficient Data Loaders for Deep Learning}},
   howpublished={\url{https://mxnet.apache.org/api/architecture/note_data_loading}},
   year={2018}
}

@misc{mlperf-results-v7,
   author={{MLPerf Training v0.7 Results}},
   title={{Designing Efficient Data Loaders for Deep Learning}},
   howpublished={\url{https://mlperf.org/training-results-0-7/}},
   year={2020}
}

@article{seda,
  title={SEDA: an architecture for well-conditioned, scalable internet services},
  author={Welsh, Matt and Culler, David and Brewer, Eric},
  journal={ACM SIGOPS Operating Systems Review},
  volume={35},
  number={5},
  pages={230--243},
  year={2001},
  %publisher={ACM New York, NY, USA}
}

@misc{cubuk2019randaugment,
    title={RandAugment: Practical automated data augmentation with a reduced search space},
    author={Ekin D. Cubuk and Barret Zoph and Jonathon Shlens and Quoc V. Le},
    year={2019},
    eprint={1909.13719},
    archivePrefix={arXiv},
    primaryClass={cs.CV}
}

@inproceedings{plinq,
author = {Tan, Roy Patrick and Nagpal, Pooja and Miller, Shaun},
title = {Automated Black Box Testing Tool for a Parallel Programming Library},
year = {2009},
%isbn = {9780769536019},
publisher = {IEEE Computer Society},
%address = {USA},
%url = {https://doi.org/10.1109/ICST.2009.32},
%doi = {10.1109/ICST.2009.32},
booktitle = {Proceedings of ICST},
pages = {307--316},
numpages = {10},
%series = {ICST ’09}
}

@inproceedings{dandelion,
author = {Rossbach, Christopher J. and Yu, Yuan and Currey, Jon and Martin, Jean-Philippe and Fetterly, Dennis},
title = {Dandelion: A Compiler and Runtime for Heterogeneous Systems},
year = {2013},
%isbn = {9781450323888},
%publisher = {Association for Computing Machinery},
%address = {New York, NY, USA},
%url = {https://doi.org/10.1145/2517349.2522715},
%doi = {10.1145/2517349.2522715},
booktitle = {Proceedings of SOSP},
pages = {49--68},
numpages = {20},
%location = {Farminton, Pennsylvania},
%series = {SOSP ’13}
}

@inproceedings{ptask,
author = {Rossbach, Christopher J. and Currey, Jon and Silberstein, Mark and Ray, Baishakhi and Witchel, Emmett},
title = {PTask: Operating System Abstractions to Manage GPUs as Compute Devices},
year = {2011},
%isbn = {9781450309776},
%publisher = {Association for Computing Machinery},
%address = {New York, NY, USA},
%url = {https://doi.org/10.1145/2043556.2043579},
%doi = {10.1145/2043556.2043579},
booktitle = {Proceedings of SOSP},
pages = {233--248},
numpages = {16},
keywords = {gestural interface, OS design, dataflow, GPUs, operating systems, accelerators, GPGPU},
%location = {Cascais, Portugal},
%series = {SOSP ’11}
}

@inproceedings{karanasos2019extending,
    title={Extending Relational Query Processing with ML Inference},
    author={Konstantinos Karanasos and Matteo Interlandi and Doris Xin and Fotis Psallidas and Rathijit Sen and Kwanghyun Park and Ivan Popivanov and Supun Nakandal and Subru Krishnan and Markus Weimer and Yuan Yu and Raghu Ramakrishnan and Carlo Curino},
    year={2020},
    booktitle={Proceedings of CIDR},
}

@article{db2,
  title={IBM Database 2 overview},
  author={Haderle, Donald J. and Jackson, Robert D.},
  journal={IBM Systems Journal},
  volume={23},
  number={2},
  pages={112--125},
  year={1984},
  %publisher={IBM}
}

@inproceedings{tensorflow,
title	= {{TensorFlow: A system for large-scale machine learning}},
author	= {Martin Abadi and Paul Barham and Jianmin Chen and Zhifeng Chen and Andy Davis and Jeffrey Dean and Matthieu Devin and Sanjay Ghemawat and Geoffrey Irving and Michael Isard and Manjunath Kudlur and Josh Levenberg and Rajat Monga and Sherry Moore and Derek G. Murray and Benoit Steiner and Paul Tucker and Vijay Vasudevan and Pete Warden and Martin Wicke and Yuan Yu and Xiaoqiang Zheng},
year	= {2016},
%URL	= {https://www.usenix.org/system/files/conference/osdi16/osdi16-abadi.pdf},
booktitle	= {Proceedings of OSDI},
pages	= {265--283}
}

@incollection{pytorch,
title = {{PyTorch: An Imperative Style, High-Performance Deep Learning Library}},
author = {Paszke, Adam and Gross, Sam and Massa, Francisco and Lerer, Adam and Bradbury, James and Chanan, Gregory and Killeen, Trevor and Lin, Zeming and Gimelshein, Natalia and Antiga, Luca and Desmaison, Alban and Kopf, Andreas and Yang, Edward and DeVito, Zachary and Raison, Martin and Tejani, Alykhan and Chilamkurthy, Sasank and Steiner, Benoit and Fang, Lu and Bai, Junjie and Chintala, Soumith},
booktitle = {Advances in Neural Information Processing Systems 32},
pages = {8024--8035},
year = {2019},
publisher = {Curran Associates, Inc.},
%url = {http://papers.neurips.cc/paper/9015-pytorch-an-imperative-style-high-performance-deep-learning-library.pdf}
}

@software{jax,
  author = {James Bradbury and Roy Frostig and Peter Hawkins and Matthew James Johnson and Chris Leary and Dougal Maclaurin and Skye Wanderman-Milne},
  title = {{JAX}: composable transformations of {P}ython+{N}um{P}y programs},
  url = {http://github.com/google/jax},
  version = {0.1.46},
  year = {2018},
}

@misc{distillation,
      title={Distilling the Knowledge in a Neural Network}, 
      author={Geoffrey Hinton and Oriol Vinyals and Jeff Dean},
      year={2015},
      eprint={1503.02531},
      archivePrefix={arXiv},
      primaryClass={stat.ML}
}
